\documentclass[preprint,12pt]{elsarticle}
\makeatletter
\let\c@author\relax
\makeatother

\usepackage[backend=biber,style=authoryear,citestyle=apa,maxbibnames=99]{biblatex}
\usepackage{amssymb}
\usepackage{siunitx}
\usepackage[flushleft]{threeparttable}
\usepackage{caption}
\usepackage{float}
\usepackage{subcaption}
\usepackage[table]{xcolor}
\usepackage{multirow}
\usepackage{textcomp}

\usepackage{graphicx}
\graphicspath{{figures/}}

\newcommand{\beginsupplement}{%
    \setcounter{subsection}{0} \renewcommand{\thesubsection}{S\arabic{subsection}}%
    \setcounter{table}{0} \renewcommand{\thetable}{S\arabic{table}}%
    \setcounter{figure}{0} \renewcommand{\thefigure}{S\arabic{figure}}%
}

\journal{arXiv}

\begin{document}

\definecolor{pink_col}{RGB}{255,200,200} 
\definecolor{red_col}{RGB}{240,0,0} 
\definecolor{green_col}{RGB}{0,200,0} 

\begin{frontmatter}



\title{Learning algorithms for identification of whisky using portable Raman spectroscopy}


\author[inst1,inst2,inst4]{Kwang Jun Lee\fnref{label1}} 

\affiliation[inst1]{organization={Centre of Light for Life (CLL) and Institute for Photonics and Advanced Sensing (IPAS)},
            addressline={The University of Adelaide}, 
            city={Adelaide},
            postcode={5005}, 
            state={SA},
            country={Australia}}

\author[inst1,inst2,inst4]{Alexander C. Trowbridge\fnref{label1}} 
\author[inst3]{Graham D. Bruce}
\author[inst3]{George O. Dwapanyin}
\author[inst4,inst5]{Kylie R. Dunning}
\affiliation[inst2]{organization={School of Physics, Chemistry and Earth Sciences},
            addressline={The University of Adelaide}, 
            city={Adelaide},
            postcode={5005}, 
            state={SA},
            country={Australia}}
\affiliation[inst3]{organization={School of Physics and Astronomy},
            addressline={University of St Andrews}, 
            city={St Andrews},
            postcode={KY16 9SS}, 
            state={Fife},
            country={United Kingdom}}
\affiliation[inst4]{organization={School of Biological Sciences },
            addressline={The University of Adelaide}, 
            city={Adelaide},
            postcode={5005}, 
            state={SA},
            country={Australia}}
\affiliation[inst5]{organization={Robinson Research Institute, School of Biomedicine},
            addressline={The University of Adelaide}, 
            city={Adelaide},
            postcode={5005}, 
            state={SA},
            country={Australia}}
      
\fntext[label1]{These authors contributed equally to this work and share first authorship}      
\fntext[label2]{These authors contributed equally to this work and share last authorship}
\cortext[cor1]{Corresponding author. Email address - kishan.dholakia@adelaide.edu.au}
\author[inst1,inst3,inst4]{Kishan Dholakia\fnref{label2}\corref{cor1}} 
\author[inst1,inst2,inst5]{Erik P. Schartner\fnref{label2}}

\begin{abstract}
Reliable identification of high-value products such as whisky is an increasingly important area, as issues such as brand substitution (i.e. fraudulent products) and quality control are critical to the industry. We have examined a range of machine learning algorithms and interfaced them directly with a portable Raman spectroscopy device to both identify and characterize the ethanol/methanol concentrations of commercial whisky samples. We demonstrate that machine learning models can achieve over 99\% accuracy in brand identification across twenty-eight commercial samples. To demonstrate the flexibility of this approach we utilised the same samples and algorithms to quantify ethanol concentrations, as well as measuring methanol levels in spiked whisky samples. Our machine learning techniques are then combined with a through-the-bottle method to perform spectral analysis and identification without requiring the sample to be decanted from the original container, showing the practical potential of this approach to the detection of counterfeit or adulterated spirits and other high value liquid samples.
\end{abstract}

\begin{graphicalabstract}
\centering
\includegraphics[width=\textwidth]{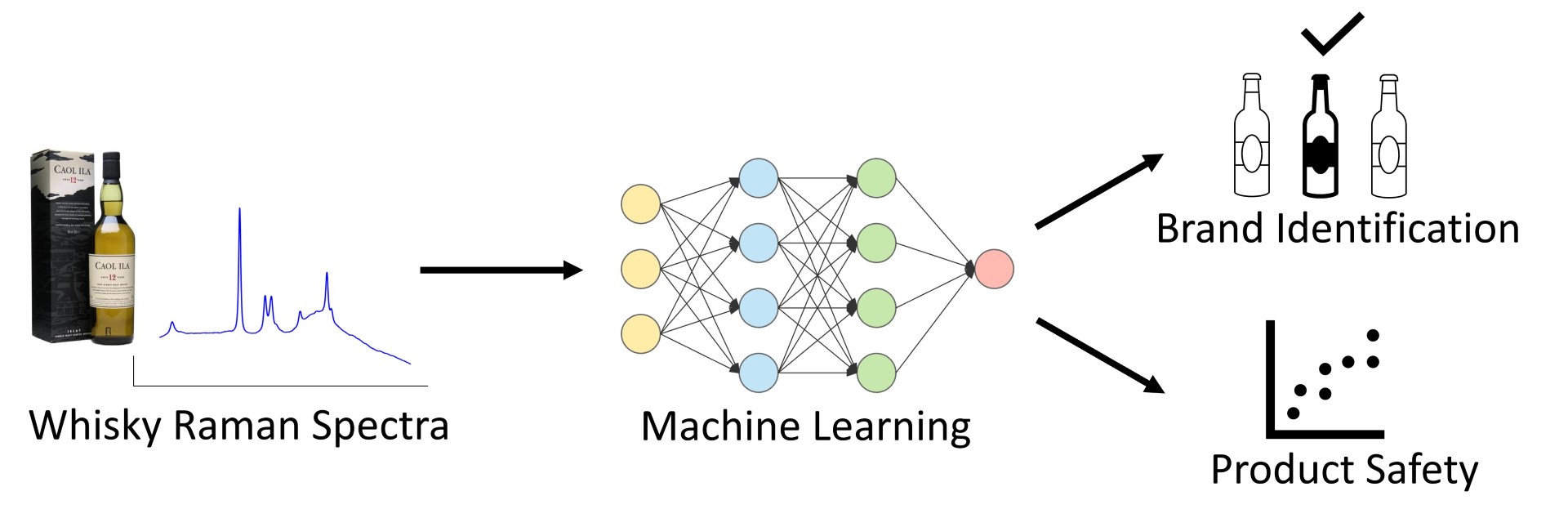}
\end{graphicalabstract}

\begin{highlights}
\item Machine learning was applied to portable Raman spectroscopy for whisky samples.
\item Trained machine learning modes predicted brand and chemical levels with high accuracy.
\item Our learning models accurately identified brands using data from unopened bottles.
\end{highlights}

\begin{keyword}
Raman spectroscopy \sep Machine learning \sep Whisky \sep Brand identification
\end{keyword}

\end{frontmatter}


\section{Introduction}
\label{sec:intro}

Recognizing a brand is important for the global growth of the whisky market as consumers are showing an increasing demand for expensive, high quality products. As demand increases, concerns arise about counterfeit and adulterated products being sold, which violate laws related to alcohol labeling and fraud (\cite{RN443,RN444}). Misrepresenting lower quality commercial whiskies as premium products can harm a producer’s reputation and financial performance. In 2018, a third of commercial Scotch whiskies
tested were found to be fraudulent  (\cite{RN445,RN446,RN447}). More broadly, the European Union loses €3 billion annually in sales due to fake wine, beer, and spirits (\cite{FAP1}). Counterfeit spirits made with industrial alcohols or poor distillation may have high levels of methanol, causing serious illness. In 2019, toxic moonshine killed 154 people in India (\cite{Toxic2019}), while in March 2020, Iranian media reported that nearly 300 people died and over 1000 became ill from drinking methanol-laced bootlegged spirits (\cite{Toxic2020}).

A range of analytical techniques, such as mass spectrometry, nuclear magnetic resonance spectroscopy, gas chromatography, and liquid chromatography, are used in laboratories to guarantee the quality, safety, and authenticity of spirits (\cite{RN442}). These techniques typically necessitate whisky sample preparation, which can be costly and time-consuming. Furthermore, the measurements cannot be performed online or continuously, and they are also not applicable to final products because one must open and remove samples under highly controlled laboratory conditions to obtain a measurement. This limits their widespread application. Portable sensors and methods for analyzing suspicious products at the point of sale or distribution are essential for widespread fraud prevention (\cite{RN453,RN452,RN450}). In this paper, a new compact Raman technique which employs machine learning is proposed to fulfill this burgeoning need.


Raman spectroscopy is a non-destructive analytical technique that uses laser light to excite the molecules in a sample and measure the inelastically scattered light. This technique can offer both quantitative and qualitative analysis of whisky samples (\cite{RN463,RN465}). It has been demonstrated to be a very versatile technique with a wide variety of uses (\cite{RN627}). This technique is not only capable of classifying whisky brands (\cite{fleming2020through,ashok2011near}), but it also has the potential to estimate the concentrations of methanol (indicative of toxicity) and ethanol (a measure of quality) in alcoholic beverages (\cite{ashok2013optofluidic}). Portable Raman spectroscopy allows for in-field analysis of whisky samples, which can be useful for monitoring the authenticity and quality of the product during production and distribution (\cite{RN460,RN459,RN455}). Importantly, techniques that can measure the Raman spectra of the contents of a bottle without opening it are also being developed (\cite{fleming2020through, shillito2022focus}).

The large amount of information from Raman spectroscopy may be efficiently handled by the use of machine learning models in the data analysis to extract valuable and subtle insights which can then be leveraged to make predictions. Standard statistical methods can struggle to fit subtle trends in complex, information rich data (\cite{RN634}). Statistical methods applying Raman spectroscopy have been already published (\cite{lednev2012raman,wang2020chemical}), suggesting the possibility to discriminate and analyse whiskies. However, a systematic comparison of different machine learning methods has not been performed for the application of Raman analysis to the brand identification of a variety of whiskies.


Recently, advancements in machine learning models have provided exciting new avenues for spectral data analysis in materials science (\cite{RN467}). Machine learning algorithms can analyze features and correlations within spectra, leading to various applications in Raman spectral analysis. Previous studies have employed various machine learning algorithms such as support vector machines (SVM), \textit{k}-nearest neighbors (KNN), and random forest (RF) (\cite{RN468,RN469}). These algorithms have demonstrated high prediction accuracy, but they also have limitations such as poor model flexibility and generalisation to new datasets, which can limit their applications in more challenging scenarios (\cite{RN475,RN470,RN471}). In particular, previous studies using these methods have required manual preprocessing to enhance the performance of machine learning in spectral data analysis. Clearly, this is a time consuming step which must be performed for each and every application (\cite{singh2021diagnosing}). Furthermore, the improper usage of preprocessing methods may lead to errors and loss of information, thereby negatively affecting the accuracy of the results and making the analysis process more complex (\cite{RN484}).


Deep learning is a subset of machine learning that is designed to handle large amounts of complex data, and can be employed to automatically extract complex features and relationships between features and tasks. Previously it has been demonstrated that deep learning can outperform conventional machine learning methods in  a range of challenging problems (\cite{sarker2021deep}),  however this is typically at the cost of higher computational resource and data requirements compared to conventional machine learning (\cite{janiesch2021machine}). 

Convolutional neural networks (CNN) are a variant of  deep learning models, and  have been successfully applied in Raman spectroscopy for the component identification of complex mixture materials (\cite{RN491,RN492}). Deep learning networks offer an advantage as they do not require manual tuning and can be trained as end-to-end networks that handle both feature extraction and classification or regression. This reduces the need for separate preprocessing or feature engineering steps, as the network can automatically convert the features into a more advanced representation (\cite{RN493}). As a result, deep learning networks can accommodate variations in samples that were previously unknown. Conventional machine learning methods, on the other hand, that require a more rigid protocol, may be unable to handle unseen data and result in inaccurate measurements. Therefore, in situations where the samples are not well-characterized or have a high degree of variability, more flexible analysis methods, such as deep learning, may be preferred. However, it should be noted that skilled preprocessing of the input data can still greatly assist the success of such models depending on the subject matter and dataset (\cite{chollet2017deep}). 



Here we focus on methods that do not require spectral preprocessing, such as smoothing, baseline correction, normalization, and spectral windowing, because it is resource intensive to perform. It can be difficult to reproduce the same results if the preprocessing steps are not well documented. In addition, spectral preprocessing can introduce bias into the data if the person performing the preprocessing is not blind to the experimental conditions. Extensive spectral preprocessing procedures must be tried to enhance outcomes. Here we used principal component analysis (PCA) rather than spectral preprocessing for our analysis of the spectral data. This choice was made because PCA is an unsupervised method that can be relatively robust and less resource-intensive compared to spectral preprocessing.

\begin{figure}[ht!]
    \centering
    \includegraphics[width=\textwidth]{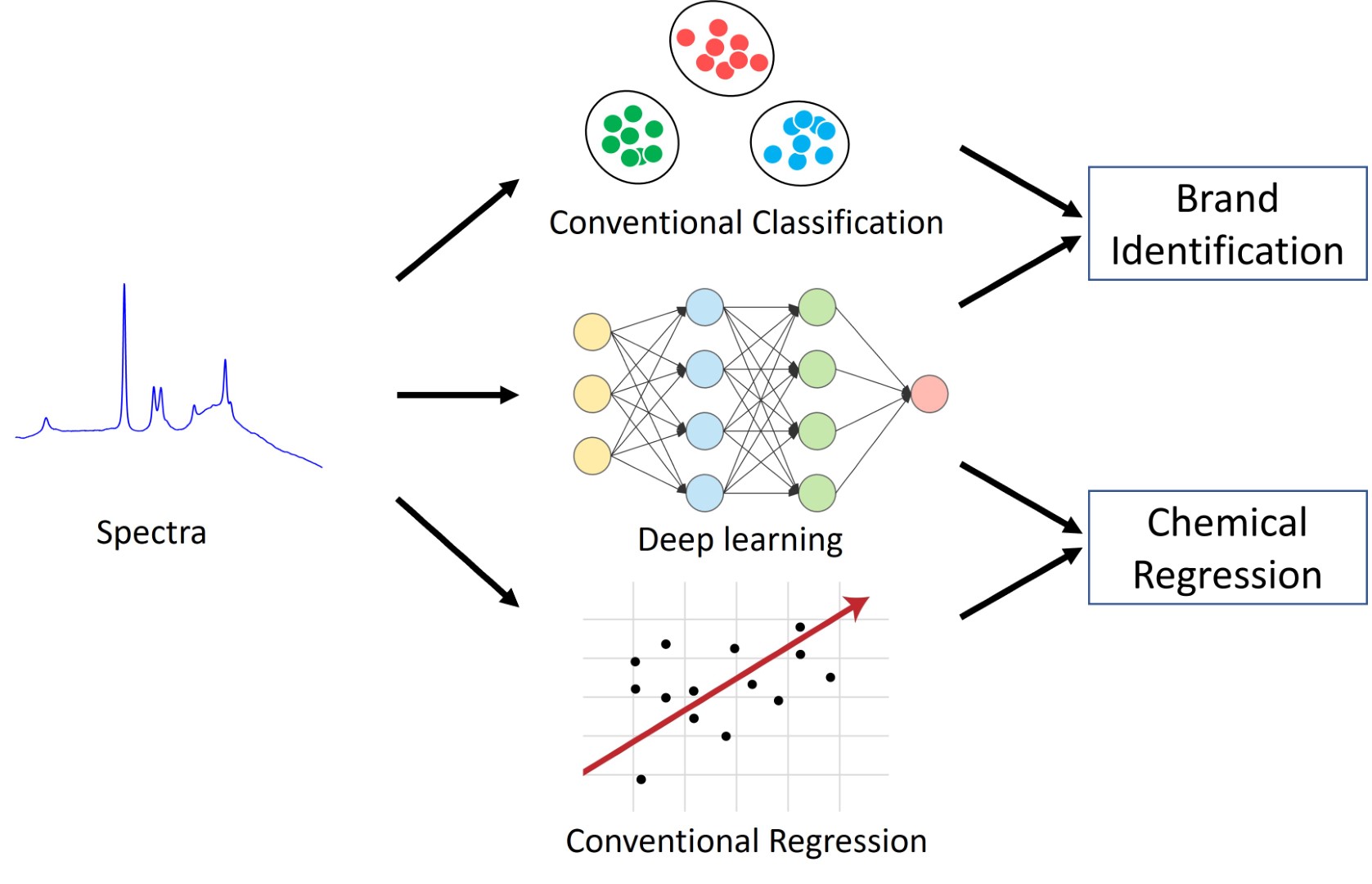}
    \caption{Overview of machine learning approaches. The figure shows the use of conventional classification and regression for brand identification and chemical regression, respectively. It also shows the use of deep learning for both brand identification and chemical regression.}
    \label{fig:method}
\end{figure}

In this study, we investigated whisky brand identification and determination of both the ethanol and methanol concentrations of whisky samples. This work provides a systematic comparison of different machine learning methods without manual spectral preprocessing for application to the brand identification of a variety of whiskies. Deep learning results were compared to conventional techniques. The machine learning methods used in this study are summarized in Figure \ref{fig:method}. We achieved over 99\% accuracy in identifying brands, and the detection uncertainty of the ethanol and methanol levels were 2.47\% volume/volume (v/v) and 0.05\% v/v, respectively. Finally, we utilized machine learning techniques to analyze the Raman spectra of the contents of a bottle without opening it. The accuracy of predicting the brand remained high even when the method was applied to datasets obtained through the bottle. 

\section{Material and Methods}
\label{sec:methods}
\subsection{Sample preparation}
A total of 28 commercially obtained whisky samples, and a reference sample of 40\% ethanol in distilled water were used for classification and ethanol quantification, as summarized in Table \ref{tbl:whisky_samples}. The whiskies were chosen to represent a variety of distilleries, flavours, cask types, and ages. The ethanol content ranges from 40\% to 63\% vol. To assess how well the trained ethanol regression models could generalize, additional test samples of twenty whiskies and three gins were used. Two pure whiskies (Talisker and Cragganmore) and a sample of 40\% ethanol/water were spiked with HPLC grade methanol. Methanol  concentrations ranging from 0-3\% in 0.3\% increments were used to generate the training set for methanol quantification. Additional test samples with 0, 0.3, 1, and 2\% methanol concentrations were prepared using Caol Ila and Cynelish via the same protocol and used to evaluate the generalization performance of the trained methanol regression models. 


\subsection{Raman analysis}
\subsubsection{Through-vial}
\begin{figure}[ht]
    \centering
    \begin{subfigure}{0.49\textwidth}
        \caption{}
        \centering
        \includegraphics[width=1\textwidth]{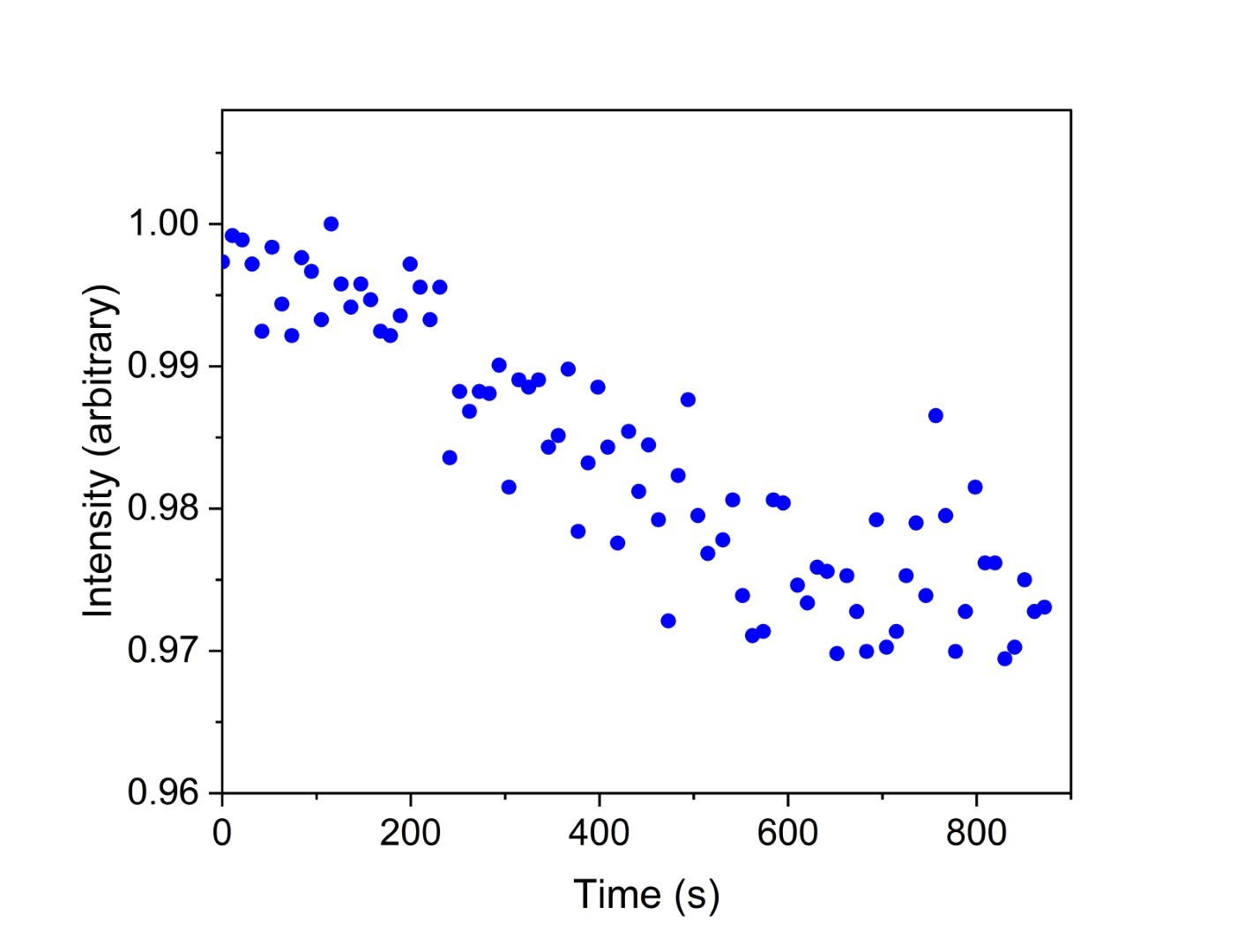} 
        \label{fig:bleaching}
    \end{subfigure}
    \begin{subfigure}{0.49\textwidth}
        \caption{}
        \centering
        \includegraphics[width=1\textwidth]{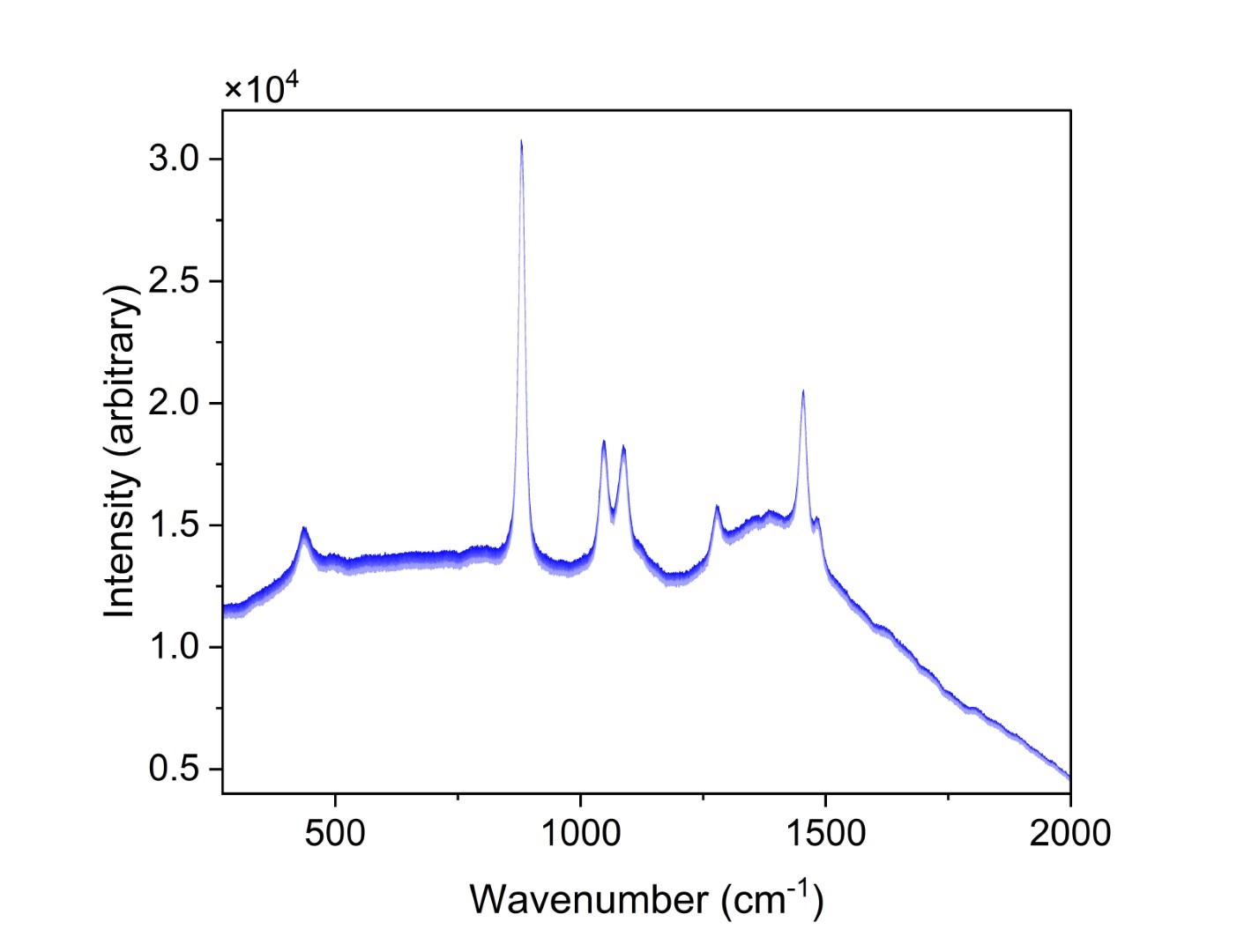}
        \label{fig:bleaching_spec}
    \end{subfigure}
    \caption{(a) Normalised intensity of the Raman signal. A series of the signal intensity at 1246 $\si{cm^{-1}}$ was recorded in Talisker during a period of 900 s and the intensity was normalised with respect to the maximum intensity. (b) All Raman spectra for 900s. The spectra are presented in shades of blue, with the darkest blue representing the initial measurement and the faintest blue representing the last measurement.}
    \label{fig:bleach}
\end{figure}

Initial studies were performed on 2 mL samples of whiskies which were pipetted from bottles and placed into 4 mL borosilicate glass vials for interrogation. All through-vial Raman spectra were collected using a compact Wasatch Photonics WP 785 Raman spectrometer (WP-785-R-SR-LMMFC-IC) using an integrated $\SI{785}{nm}$ laser and $\SI{25}{\micro m}$ slit giving a resolution of $\SI{7}{cm^{-1}}$. The laser was coupled into a Raman probe, and the emission collected by the probe was transferred to the spectrometer. The probe was focused directly into a clear glass vial containing the whisky. All spectra were collected in the spectral range $\numrange[range-phrase=-]{270}{2000} \si{cm^{-1}}$ with a laser power of 450 mW, an integration time of 500 ms and an average of 5 scans. Each whisky sample had forty replicates collected, except for the methanol test samples which had twenty replicates collected. Continuous irradiation for 15 minutes resulted in a 3\% reduction in the intensity of spectra, as can be seen in Figure \ref{fig:bleach}. During the measurement process, each sample was irradiated for about 1 minute. This short exposure time had a negligible effect on photo-bleaching, as it only resulted in a 0.2\% reduction in intensity.   

\subsubsection{Through-bottle}
\begin{figure}[ht!]
    \centering
    \includegraphics[width=\textwidth]{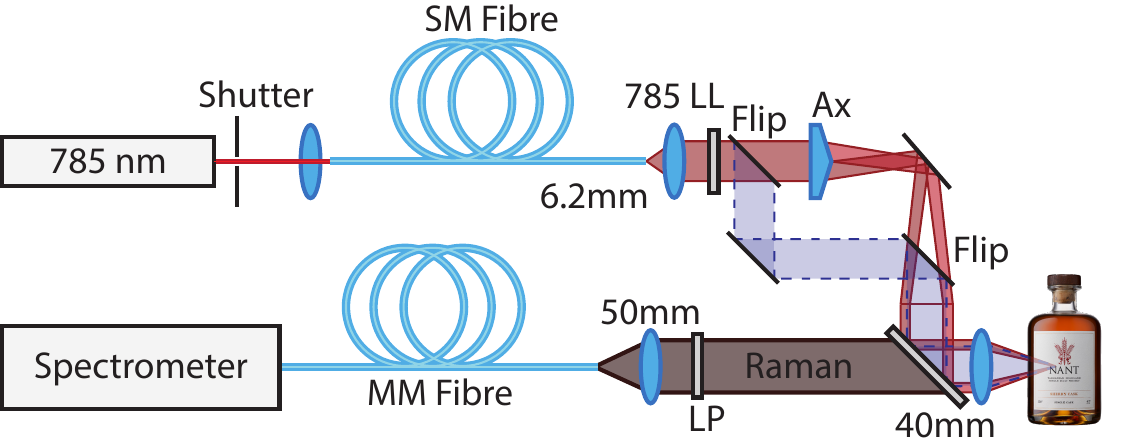}
    \caption{Experimental configuration for through the bottle Raman measurements. Red path shows the Bessel Through the bottle beam, blue shows conventional backscattering (gaussian profile), and brown the collected Raman signal. 785 LL - Semrock LL01-785-12.5, Ax - Thorlabs AX255-B, DC - Semrock LPD02-785RU-25x36x1.1, LP - Semrock LP02-785RU-25, Flip - Mirrors mounted on adjustable mounts to swap between the two beam paths}
    \label{fig:TTB}
\end{figure}

Through the bottle Raman measurements were performed using a free-space system, consisting of a Spectra-Physics 3900s Ti:Sapphire tunable laser for excitation and an Andor Shamrock SR-303i spectrometer for spectra collection. The experimental setup in this section was based on that previously demonstrated in \cite{fleming2020through} and \cite{shillito2022focus}, and utilized an axicon-based focus-matched inverse spatially-offset Raman configuration. Two configurations were set up to allow switching between the axicon configuration and a conventional back-scattering (Gaussian beam profile) configuration through the use of flip mirrors. The system was  aligned such that the two paths were collinear with the focal point optimized for maximum Raman signal collection as shown in Figure \ref{fig:TTB}. All spectra were collected in the spectral range $\numrange[range-phrase=-]{140}{2700} \si{cm^{-1}}$ with a laser power at the sample of 96 mW for the Gaussian beam and 105 mW for the Bessel, and an integration time of 5 seconds. Each whisky sample (see Table \ref{tbl:whisky_samples_ttb} for details) had 30 replicates collected. As for the through-vial analysis, a measurement was performed every second for 60 minutes to observe if the signal from the Whisky reduced over time due to photobleaching. Results here demonstrated that the signal remained within \textpm 2\% of the initial peak value.

To thoroughly examine the performance of the machine learning methods, spectra were collected with this system both in the original (thick glass) bottles, as well as decanting the samples into thin-walled 4 mL vials similar to those used in the through-vial measurements.

\subsection{Data processing}
Deep learning networks were performed using Python 3.9.15, sci-kit learn 1.1.3 and TensorFlow 2.11.0 on a computer equipped with an NVIDIA GeForce RTX 4090 GPU and an Intel Core i9-12900KS CPU. Conventional classification machine learning methods were performed using Python 3.11.3 and sci-kit learn 1.2.2 on a computer equipped with an NVIDIA GeForce GTX 1650 Ti mobile GPU and an AMD Ryzen 7 4800HS mobile CPU. PCR, PLSR, and ridge regression were processed using MATLAB R2021b software (MathWorks, Natick, USA) on a computer equipped with an NVIDIA GeForce GTX 1650 Ti mobile GPU and an AMD Ryzen 7 4800HS mobile CPU.

\subsubsection{Deep learning models}
\begin{figure}[ht]
    \centering
    \includegraphics[width=\textwidth]{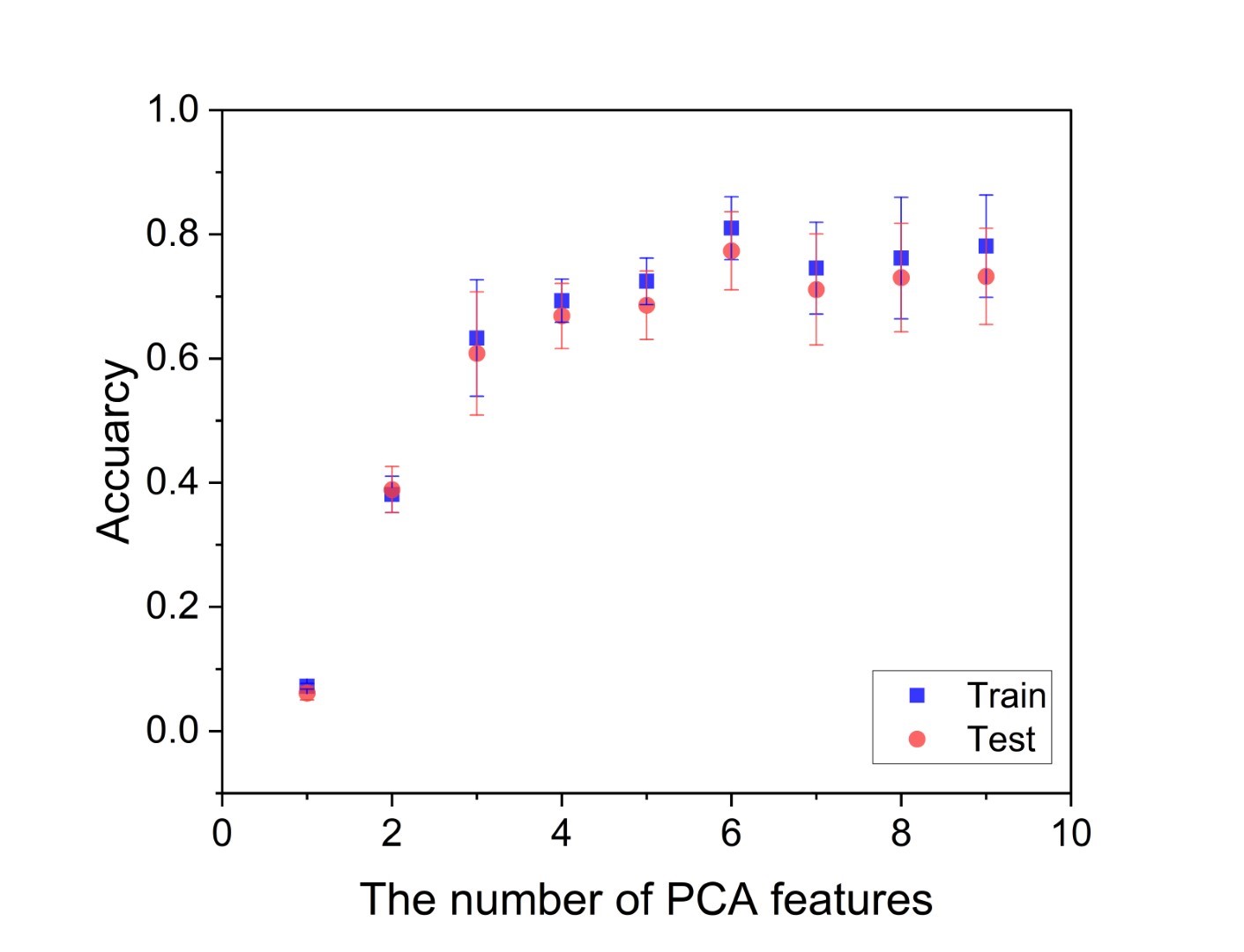}
    \caption{Brand identification accuracy of ANN model for different number of features. The blue squre and red dots represent the results of the training and test sets, respectively, depending on PCA features. Error bars represent standard deviations of six repeats.}
    \label{fig:ANN}
\end{figure}

Three different configurations of the deep learning model, including CNN, fully connected networks (FCN), and a hybrid parallel model (HPM), which is a combination of CNN and FCN, were applied to predict brand identification, ethanol concentration, and methanol concentration using the same analysis model. The spectral data were split into training (60\%), validation (20\%), and test (20\%) sets for brand identification and ethanol quantification. The spectral data of Talisker, Cragganmore, and 40\% ethanol/water spiked with methanol were divided into training (80\%) and validation (20\%) sets, while the spectra of Caol Ila and Clynelish spiked with methanol were used exclusively as a test set for methanol quantification as summarized in Table \ref{tbl:Methanol dataset}. The raw spectral data was either directly used as the input for deep learning models or the spectral data underwent PCA, and six PCA features were selected based on preliminary results summarised in Figure \ref{fig:ANN}. The performance of the ANN model, which is closely related to deep learning models, reached a maximum at six PCA features. In this case the artificial neural network (ANN) model has only a few layers and is hence not considered to be \emph{deep} learning. As a result, it was determined that six was the best number of components to use. These selected features served as the input for the classification and regression process. The prediction accuracy was employed to evaluate the brand identification performance of different methods. The root mean square error (RMSE) and determination coefficient ($R^2$) were used to evaluate the quality of quantification analysis. RMSE and $R^2$ of the training, validation, and test sets are abbreviated as RMSE\textsubscript{T} and $R^2_T$, RMSE\textsubscript{V} and $R^2_V$, and RMSE\textsubscript{P} and $R^2_P$, respectively. The values of parameters for deep learning models are described in Table \ref{tbl:DL_parameters}.

\subsubsection{Conventional classification machine learning}
The spectral data was divided into training (70\%) and test (30\%) sets for conventional classification machine learning. In this study, the unprocessed spectral data was either directly used as input for conventional machine learning models, or subjected to PCA which was used to reduce the dimensionality of the data and also provided an initial evaluation of the data's predictability (\cite{pca2014tutorial}).

The raw spectral data was either directly used as the input for conventional machine learning models or the spectral data underwent PCA, and from one to nine PCA features were used as input for classification. PCA is an unsupervised learning method, which means that it does not require data that has been assigned a class or category to train the model. The lack of supervision allows the method to be applied to any dataset, as opposed to requiring a new method to be optimized each time (\cite{pca2014tutorial}). In order to dramatically reduce the search space and minimise the associated computational time for each technique, a robust PCA model was constructed. This enables the retention of the most useful information in the data while efficiently discarding the excess. The parameter values for conventional classification machine learning are described in Table \ref{tbl:conv_ML_parameters}.

\subsubsection{Conventional regression machine learning for methanol quantification}
Three regression methods were evaluated: principal component regression (PCR), partial least squires regression (PLSR), and ridge regression. Table \ref{tbl:Methanol dataset} describes how the experimental set was divided. As a training set, spectral data from Talisker, Cragganmore, and 40\% ethanol/water spiked with methanol were used, and cross-validation was performed using venetian blinds with five cross-validation groups. The spectra of Caol Ila and Cynelish spiked with methanol were used as a test set. The values of RMSE and $R^2$ were used to evaluate the quality of quantification analysis.

\section{Results}
\subsection{Deep Learning}
\begin{figure}[ht]
    \centering
    \includegraphics[width=0.8\textwidth]{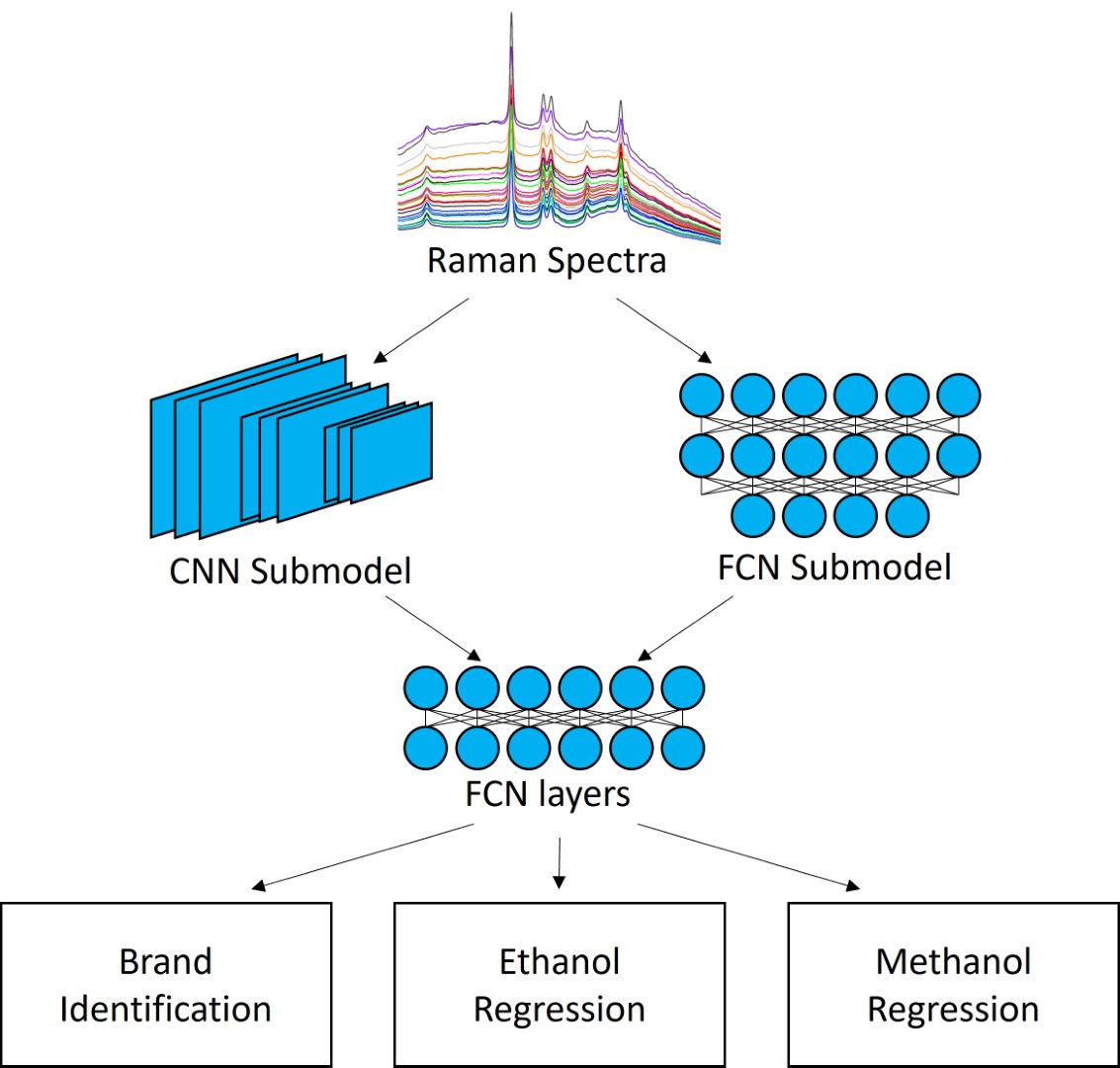}
    \caption{The hybrid model (HPM) uses the full architecture presented here. When only the CNN or FCN models are used, one of the internal branches is simply omitted.}
    \label{fig:dl_model}
\end{figure}

In this section, we evaluate the performance of different deep learning approaches in identifying brands and quantifying ethanol/methanol. The deep learning models that were used in this study are described in Figure \ref{fig:dl_model} and the supplementary material section S2.

\subsubsection{Brand identification} 
\begin{figure}[H]
    \centering
    \begin{subfigure}{0.5\textwidth}
        \caption{}
        \centering
        \includegraphics[width=1\textwidth]{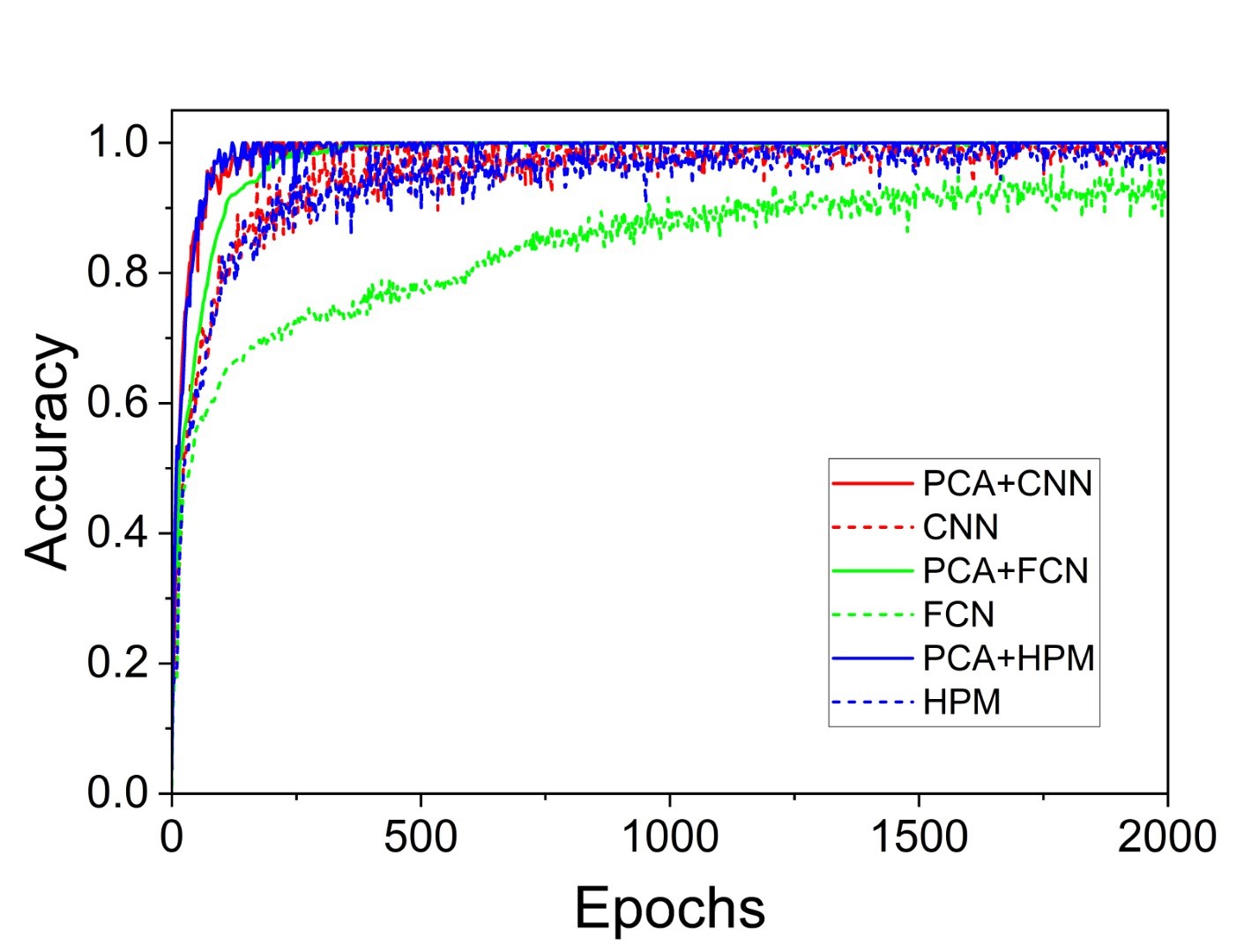} 
        \label{fig:Accuracy}
    \end{subfigure}\hfill
    \begin{subfigure}{0.5\textwidth}
        \caption{}
        \centering
        \includegraphics[width=1\textwidth]{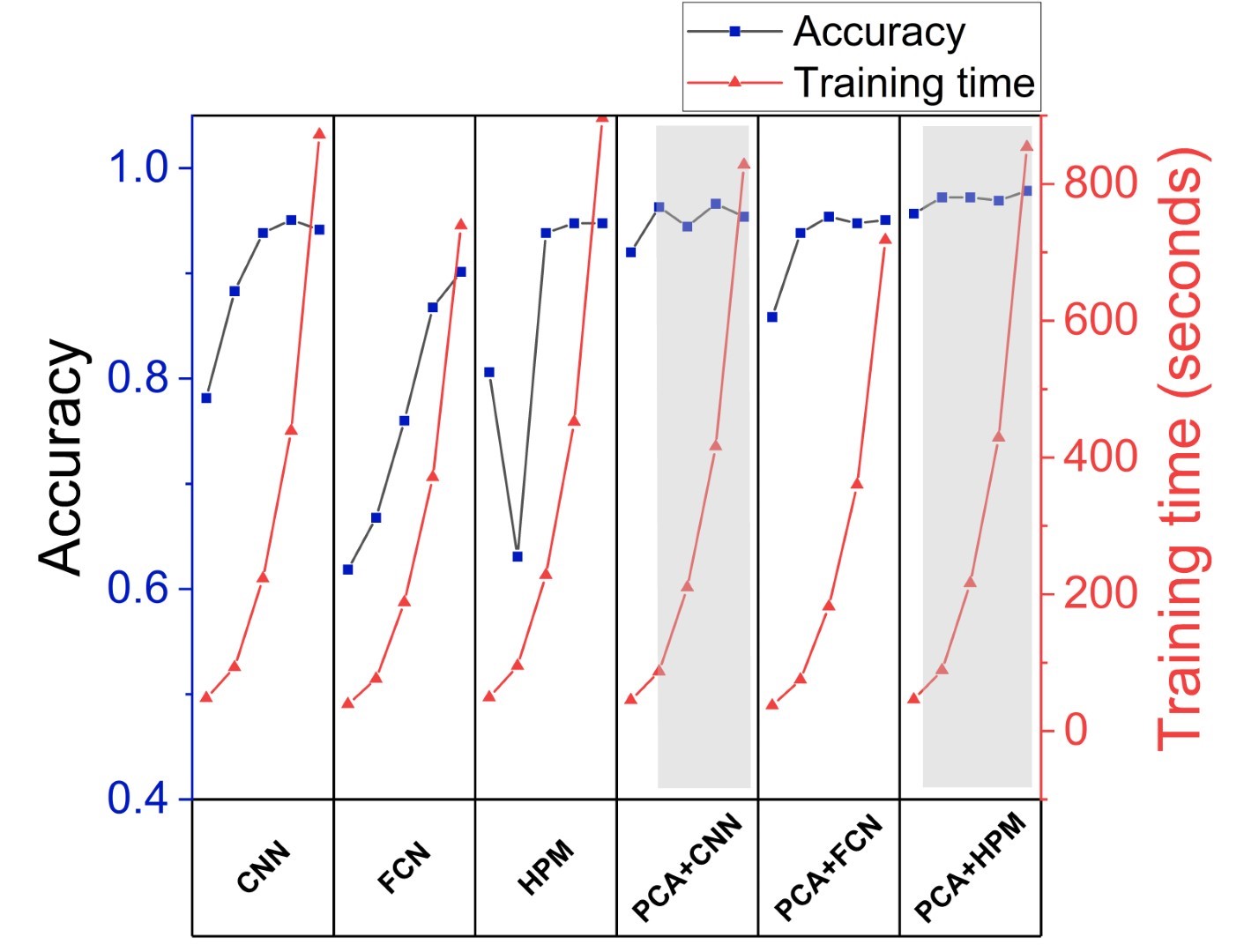} 
        \label{fig:Accuracy_Time}
    \end{subfigure}
    \caption{Brand identification: (a) Deep learning model accuracies per epoch on the training set. (b) Accuracy of deep learning model on the test set and training time. The blue and red dots represent the test accuracy and training time results, respectively. Each dot represents an increase in accuracy and training time as the epochs increase by 100, 200, 500, 1000, and 2000. Shaded grey areas show where the model has $>$ 96\% accuracy.}
        \label{fig:Identification}
\end{figure}

We compared three different deep learning methods: CNN, FCN, and HPM. PCA was also combined with each of these methods (resulting in PCA+CNN/FCN/HPM). We trained each method with different numbers of epochs (100/200/500/1000/2000). The accuracy results depending on epochs are summarized in Figure \ref{fig:Identification} and Table \ref{tbl:deeplearning_class}. All models with above 500 epochs demonstrated greater than 93\% test accuracy except for FCN without PCA. Increasing the number of epochs used for training had a larger impact on accuracy, as many models had higher accuracy when the number of epochs was increased from 200 to 2000. However, as can been seen in Figure \ref{fig:Accuracy}, all models converged to a high accuracy before oscillating depending on the data in the set. It is interesting to note that the use of PCA appeared to shorten the accuracy plateau time of CNN and HPM while also increasing the prediction accuracy of FCN from 90\% to 95\%. This suggests that reducing the dimensionality of the data using PCA before training could be an effective approach for improving the performance of deep learning models. Overall, PCA+HPM outperformed all other deep learning models on this dataset, with training times of 89 seconds, and more than 97\% accuracy on all datasets when 200 or more epochs were used (Figure \ref{fig:Accuracy_Time}). Even when trained for only 100 epochs, this model achieved a test accuracy of 96\% in just 46 seconds of training time.

\subsubsection{Chemical regression} \label{sec:chem_reg}

\begin{figure}[H]
    \centering
    \begin{subfigure}{0.45\textwidth}
        \caption{}
        \centering
        \includegraphics[width=1\textwidth]{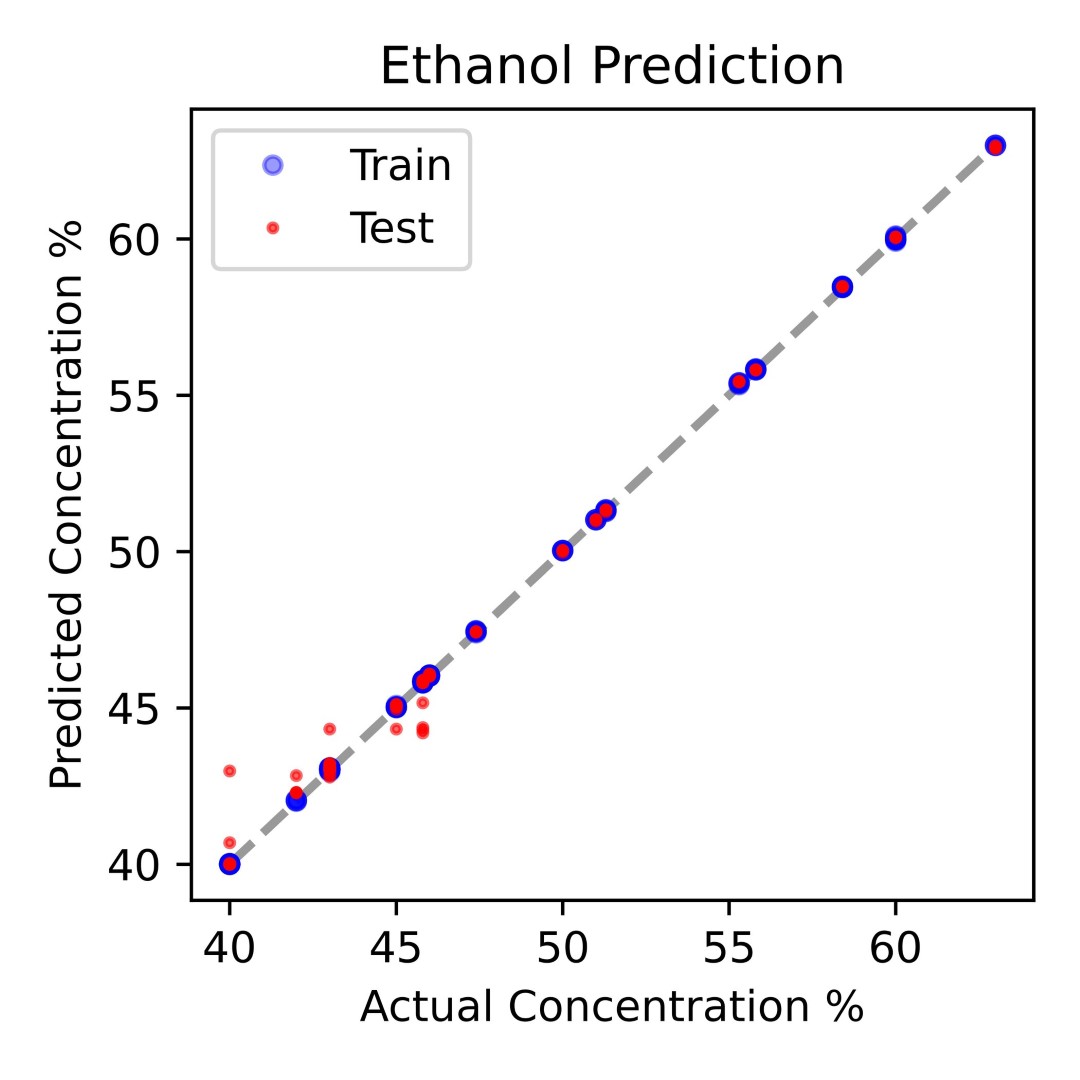} 
        \label{fig:EtOH_2000}
    \end{subfigure}\hfill
    \begin{subfigure}{0.45\textwidth}
        \caption{}
        \centering
        \includegraphics[width=1\textwidth]{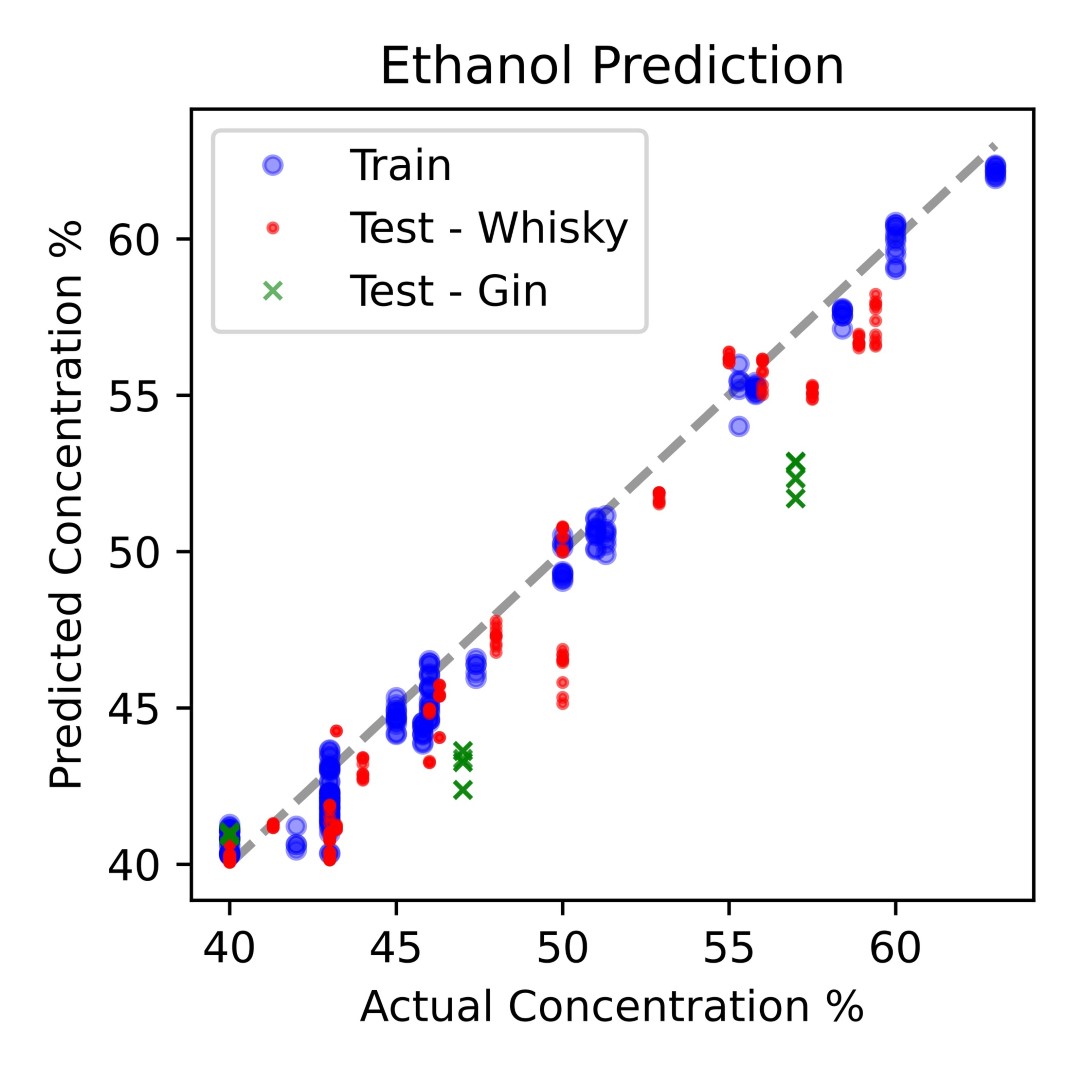} 
        \label{fig:EtOH_external}
    \end{subfigure}
    \begin{subfigure}{0.45\textwidth}
        \caption{}
        \centering
        \includegraphics[width=1\textwidth, trim={0cm 0cm 0cm 0cm}, clip]{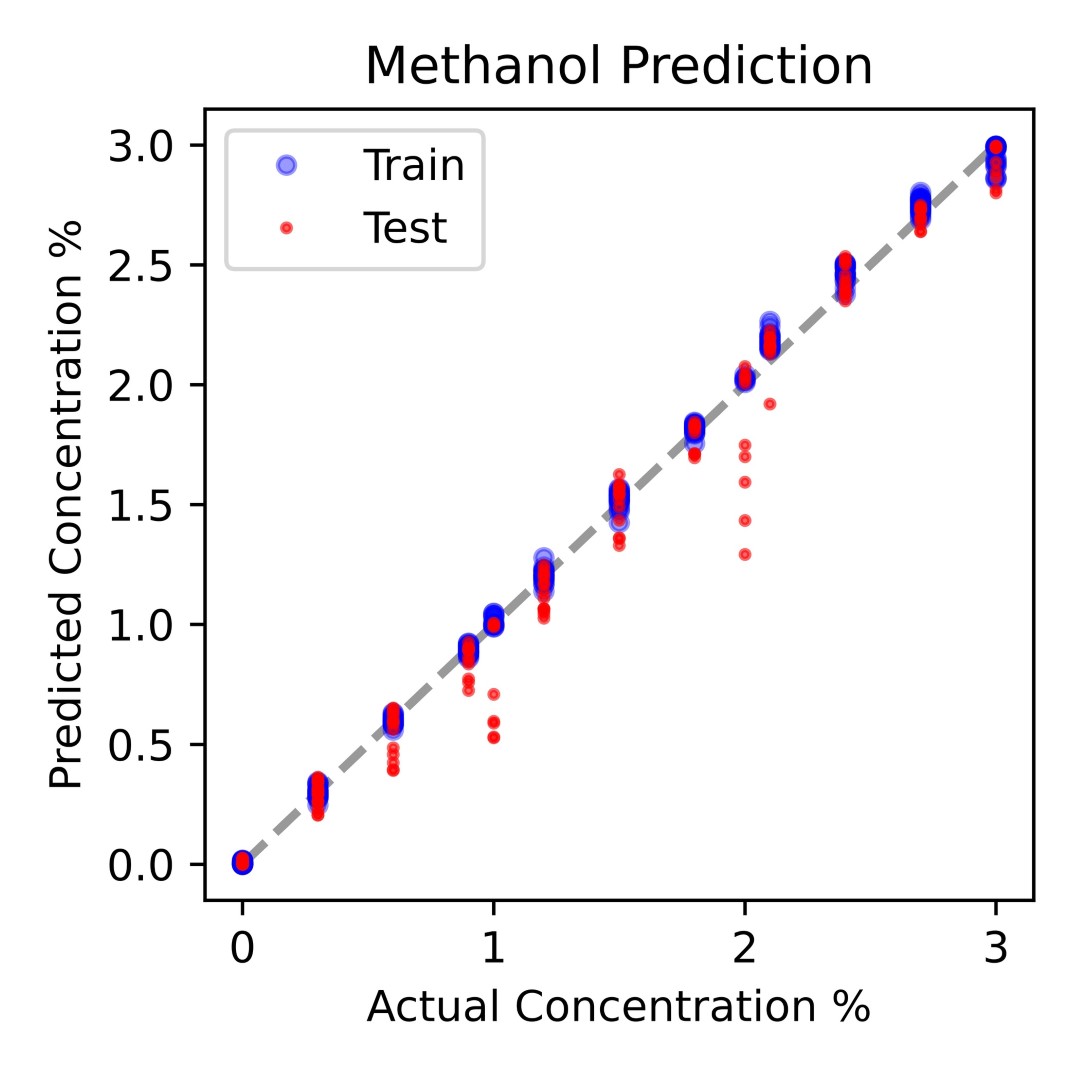} 
        \label{fig:MeOH_normal} 
    \end{subfigure}\hfill
    \begin{subfigure}{0.45\textwidth}
        \caption{}
        \centering
        \includegraphics[width=1\textwidth, trim={0cm 0cm 0cm 0cm}, clip]{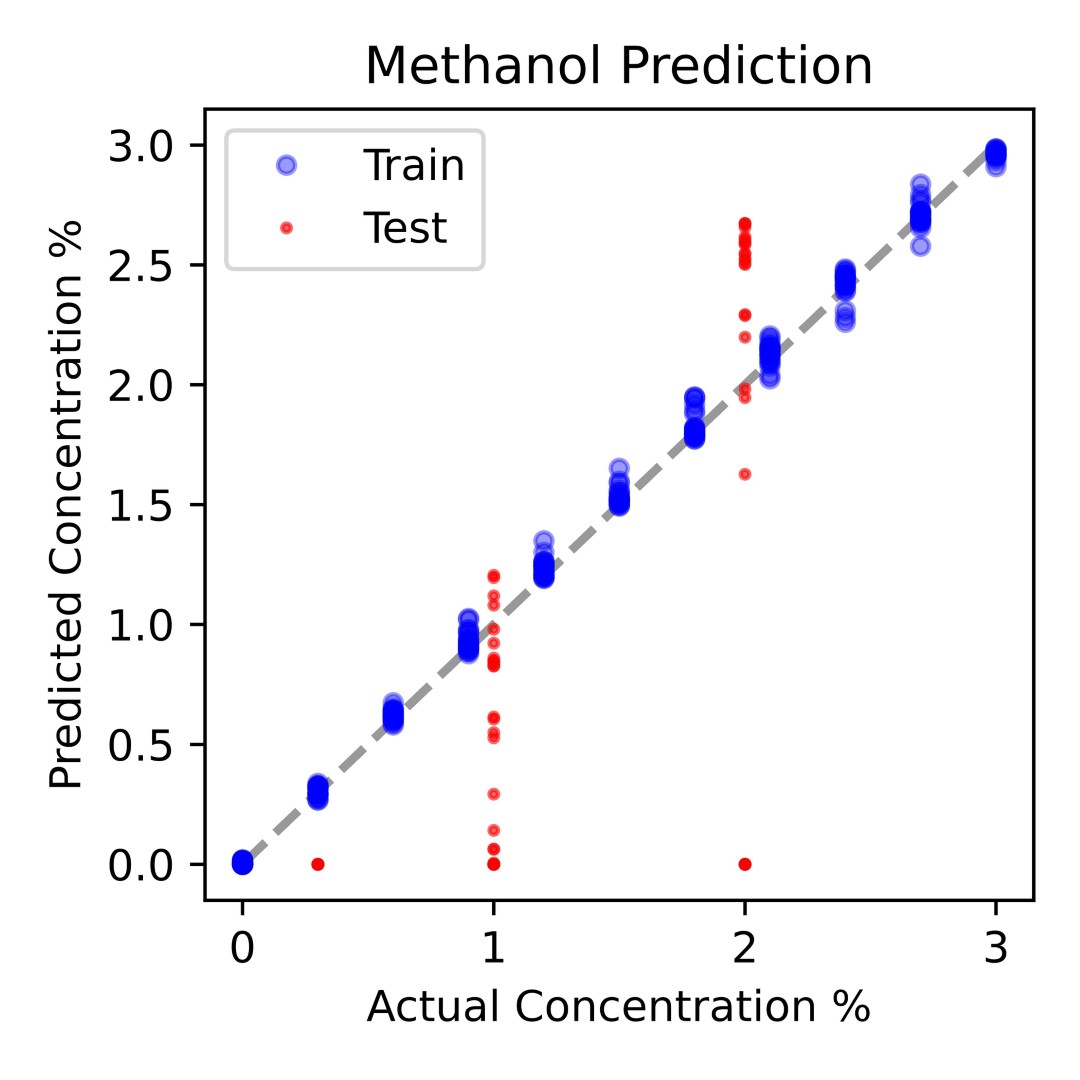} 
        \label{fig:HPM_MeOH}
    \end{subfigure}\hfill
    \caption{Ethanol content prediction: (a) Using PCA+HPM with 2000 epochs ($R^2$ = 0.998 and RMSE = 0.25\% for the test set). (b) Using PCA+FCN with 200 epochs applied to a dataset with previously unseen whisky brands (red dot) and gin samples (green x) ($R^2$ = 0.863 and RMSE = 2.47\% for the test set). Methanol content prediction: (c) Using PCA+HPM with 1000 epochs. The spectral data of Talisker, Cragganmore, 40\% ethanol/water, Caol Ila, and Cynelish spiked with methanol were divided into training (60\%), validation (20\%), and test (20\%) sets. (d) Using PCA+HPM with 1000 epochs. The spectral data of Talisker, Cragganmore, and 40\% ethanol/water spiked with methanol were split into training (80\%) and validation (20\%) sets. The spectra of Caol Ila and Cynelish spiked with methanol were only used as a test set. The blue and red dots represent the results of the training and test sets, respectively.}
    \label{fig:EtOH}
\end{figure}

The same deep learning algorithms and data sets used for brand identification were used to predict ethanol content in whisky samples. The results are summarized in Figure \ref{fig:EtOH_2000} and \ref{fig:EtOH_external}, and represented in more detail in Table \ref{tbl:deeplearning_alcohol}. Among the deep learning models without PCA, the CNN method achieved the best results with the highest $R^2$ score of 0.994 and the lowest RMSE of 0.39\% (in terms of ethanol in the sample) on the test set. The HPM method achieved competitive results compared to CNN, with the maximum $R^2$ score of 0.993 and RMSE of 0.43\% on the test set. All models with PCA performed significantly better than models without PCA. PCA+CNN and PCA+HPM displayed the best performance with a $R^2$ of 0.998 and RMSE of 0.24\% and 0.25\%, respectively, on the test set.  

The deep learning models were tested on samples that were not used during training to evaluate their ability to generalize. The unseen test set included twenty whisky and three gin samples. The best results were achieved using the PCA+FCN model when trained for 200 epochs, with an $R^2$ of 0.863 and RMSE of 2.47\% on the unseen test set (Figure \ref{fig:EtOH_external}). The performance of ethanol content prediction on the unseen test set is shown in Table \ref{tbl:deeplearning_alcohol}. It is conjectured that as the FCN model is simpler, it can more quickly fit to a particular spectrum (e.g. ethanol) whereas for brand identification the more complex CNN-based models are required. Increasing the number of epochs resulted in higher RMSE values, indicating that optimizing the number of epochs is necessary for generalizing the model. The three deep learning models without PCA did not perform well on the unseen test set. Interestingly, while the performance of the PCA+HPM model improved with an increase in the number of epochs, the performance of the PCA+CNN and PCA+FCN models decreased. Increasing the number of samples in the training set or the number of epochs may improve the performance of the PCA+HPM model on the unseen test set. The predicted ethanol content for gin samples was slightly lower than the actual content, but the model still performed well, which is notable considering that gin was not included in the training data (Table \ref{tbl:deeplearning_alcohol}). This suggests that the deep learning models were able to generalize well and accurately predict the ethanol content in samples that were not used during training.

The methanol prediction used the same deep learning algorithms as the brand identification and ethanol prediction. The results are presented in Table \ref{tbl:deeplearning_methanol}, and demonstrate the model performed poorly on the test set compared to the training and validation sets, indicating over-fitting. The PCA+HPM model showed the best performance among all the models. It achieved the lowest RMSE of 0.98\% (in terms of methanol in the sample) for the test set. However, its performance ($R^2$ value of 0.300) was still unsatisfactory. The model showed a very good fit when it trained and tested on the same samples. Figure \ref{fig:MeOH_normal} shows the performance of the model when the training set and test sets are comprized of the same whisky samples, with identical concentrations of methanol. As can be seen, the $R^2$ values were 0.999 for the training set and 0.991 for the test set. Figure \ref{fig:HPM_MeOH} shows the model results when the test set is similar whiskies from different manufacturers, with methanol concentrations not included in the training set. The model was unable to accurately predict methanol concentrations in the samples. 

\subsection{Conventional Machine Learning}
\subsubsection{Brand identification}
Nine conventional machine learning techniques were applied to classify whisky brands using Raman spectra. All of the conventional machine learning models that were used in this study are described in the supplementary material section S3.  Figure \ref{fig:convention} shows the accuracy results for conventional machine learning with and without PCA. Several machine learning algorithms including KNN, RF, and linear discriminant analysis (LDA) without using PCA achieved a high level of test accuracy, exceeding 96\%. Several machine learning methods had significantly improved performance when using PCA. In particular, the test accuracy of quadratic discriminant analysis (QDA) increased from 25\% to over 99\% when using three or more PCA features while the test accuracy of artificial neural network (ANN) improved from 4\% to 85\% when using six PCA features. These findings support that PCA can effectively reduce noise in the data and improve classifier performance. It is worth noting that although certain methods, such as radial basis function SVM and Gaussian Process, achieved high accuracy on the training set, this did not result in high accuracy on the test set, with values lower than 10\%.

\begin{figure}[H]
    \centering
    \includegraphics[width=1\textwidth]{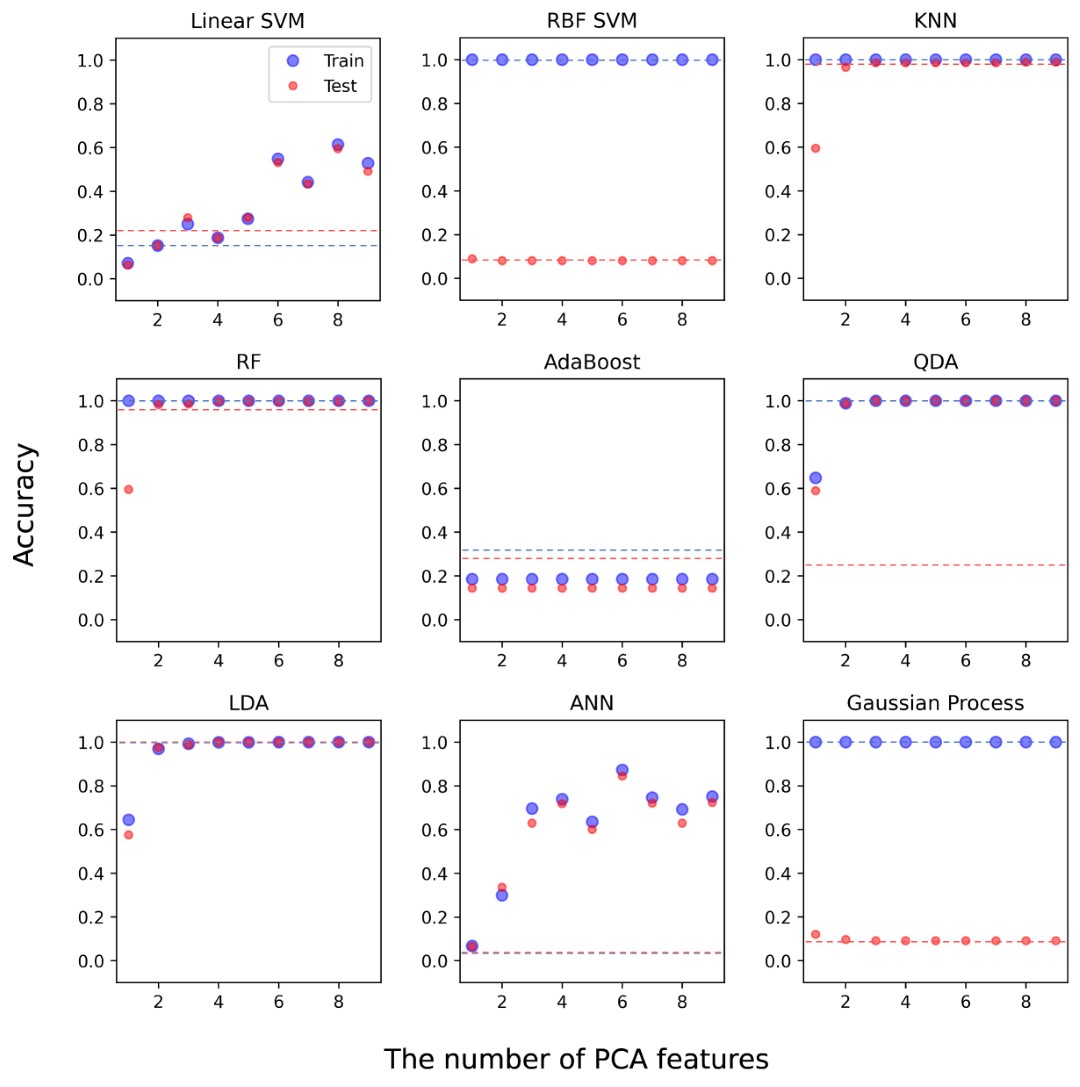}
    \caption{Brand identification accuracy of conventional machine learning models. The blue and red dots represent the results of the training and test sets, respectively, with PCA. The dashed lines show the respective accuracy levels without PCA. The red and blue dashed lines overlap each other on the LDA and ANN graphs due to the identical accuracy they achieved on both the training and test sets.}
    \label{fig:convention}
\end{figure}


\subsubsection{Methanol regression}
\begin{figure}[H]
\centering
    \includegraphics[width=0.7\textwidth]{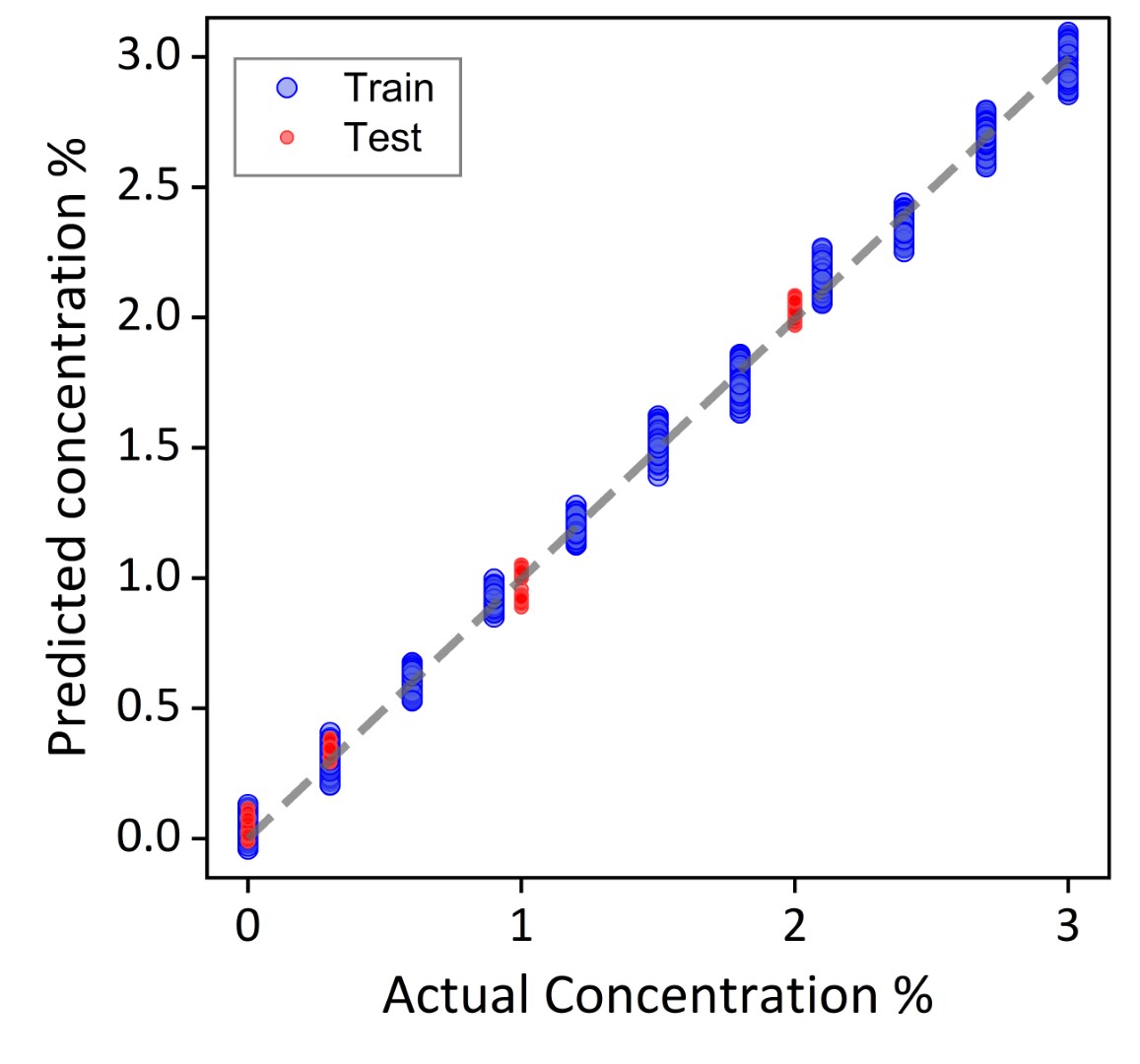} 
    \label{fig:PLSR_MeOH}
    \caption{Methanol content prediction using a PLSR model (PLSR7). The spectral data of Talisker, Cragganmore, and 40\% ethanol/water spiked with methanol were used as a training set. The spectra of Caol Ila and Cynelish spiked with methanol were only used as a test set. The blue and red dots represent the prediction results on the training and unseen test sets, respectively.}
    \label{fig:MeOH}
\end{figure}

\begin{figure}[ht]
    \centering
    \begin{subfigure}{0.33\textwidth}
        \caption{}
        \centering
        \includegraphics[width=1\textwidth]{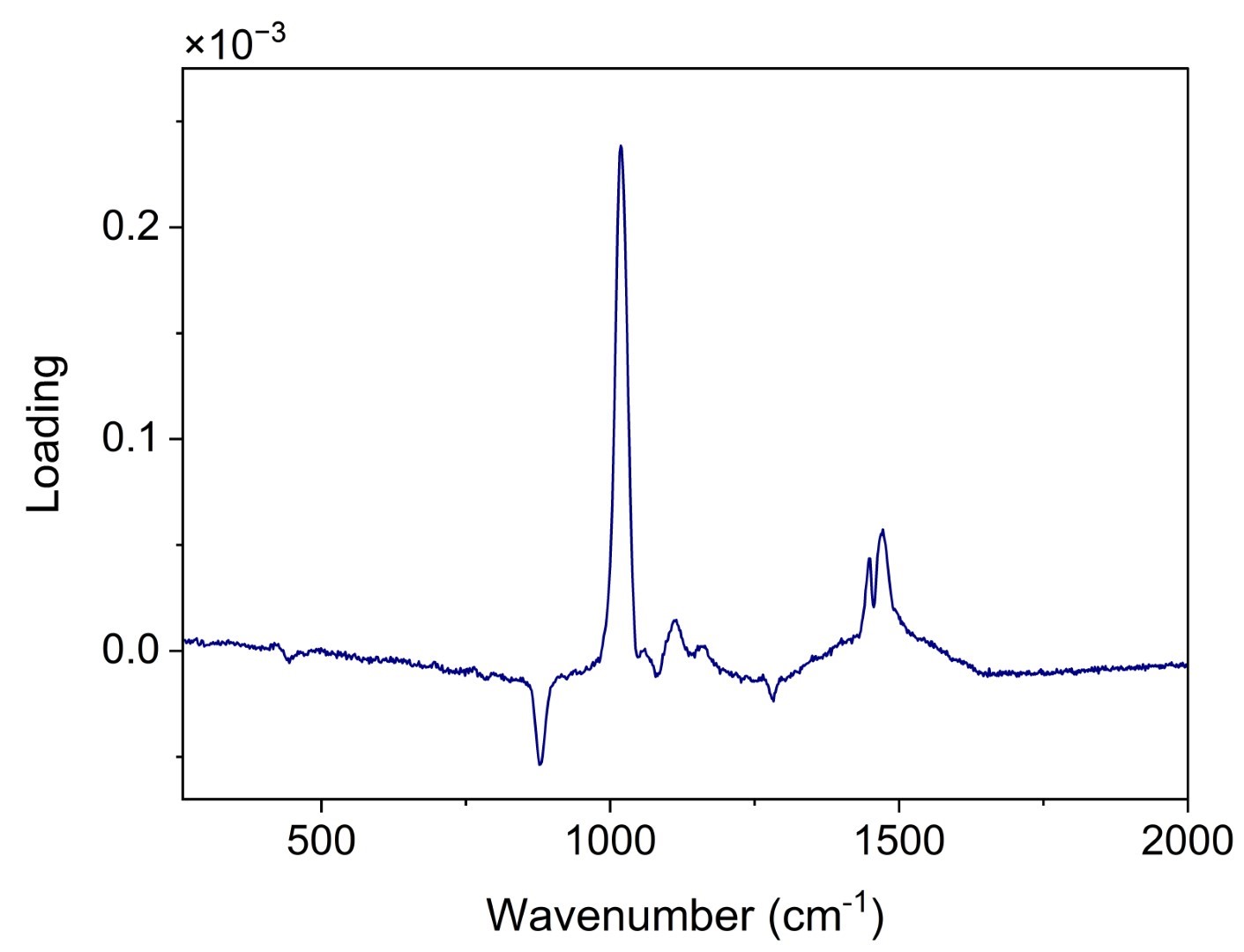} 
        \label{fig:MeOH_loading}
    \end{subfigure}\hfill
    \begin{subfigure}{0.33\textwidth}
        \caption{}
        \centering
        \includegraphics[width=1\textwidth]{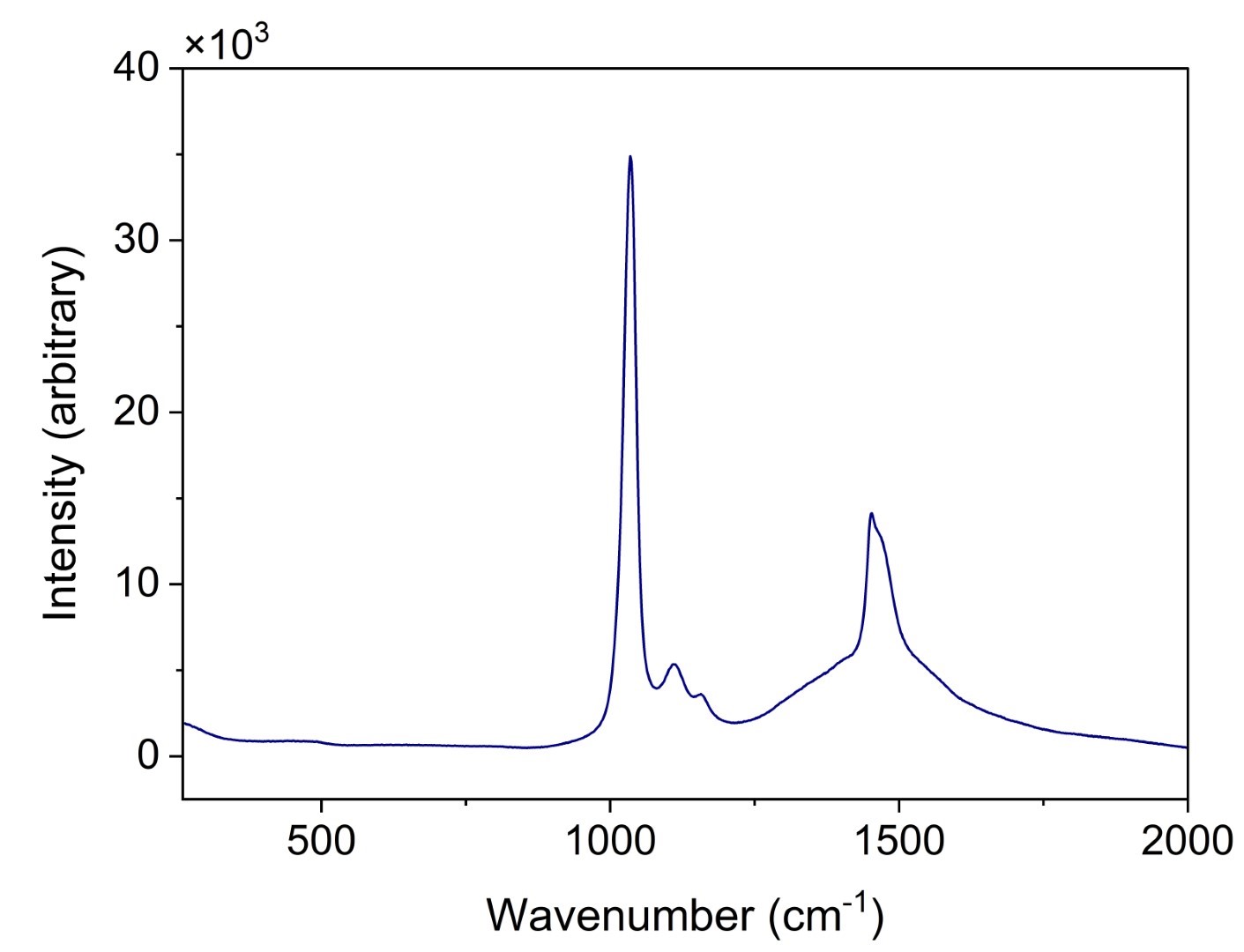} 
        \label{fig:MeOH_Raman}
    \end{subfigure}
    \begin{subfigure}{0.33\textwidth}
        \caption{}
        \centering
        \includegraphics[width=1\textwidth]{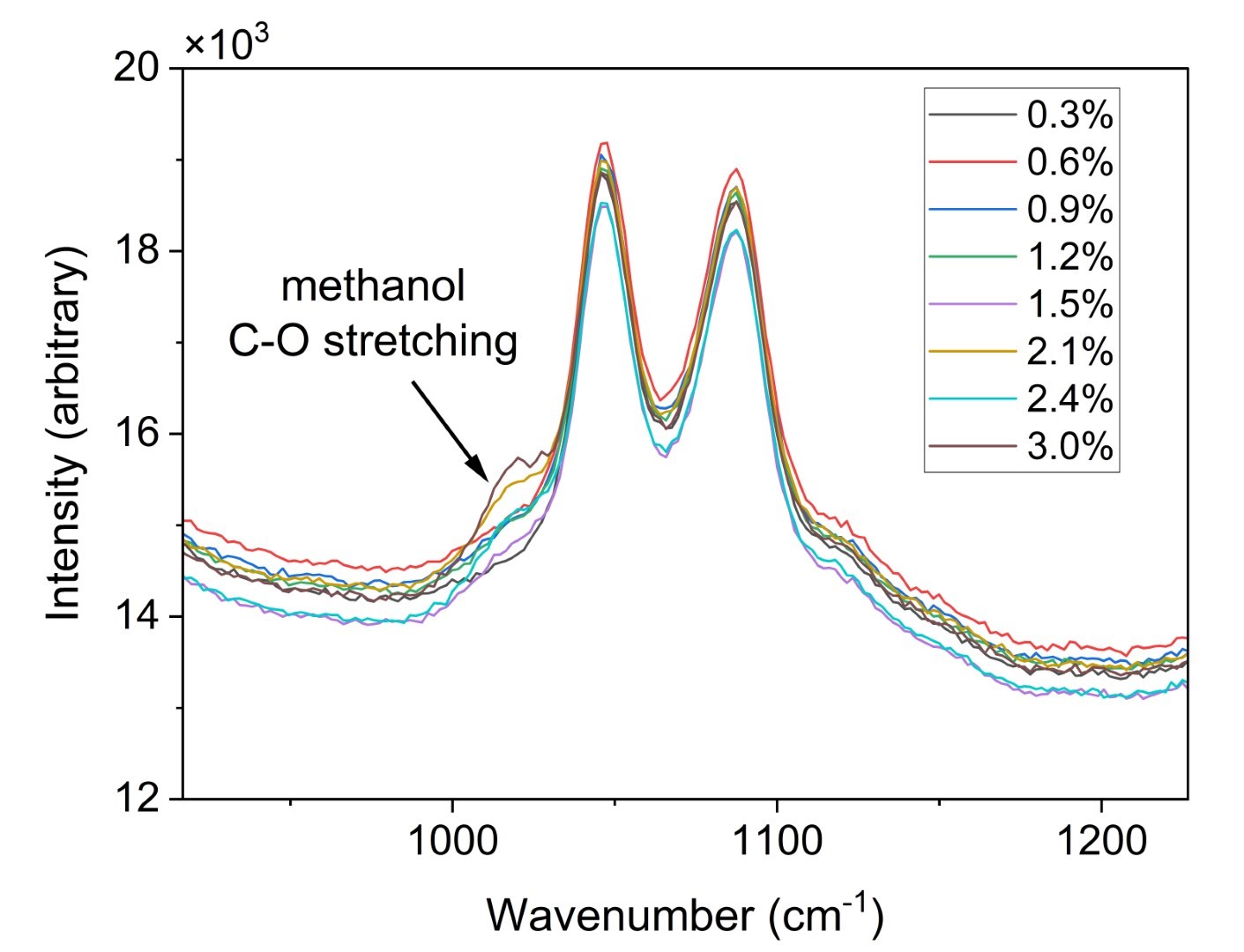}
        \label{fig:EtOH_MeOH}
    \end{subfigure}
    \caption{(a) The loading plot of the PLSR7 model, suggested the methanol vibration at near 1020 \si{cm^{-1}} as the most significant variables contributing to the model. (b) A Raman spectrum of methanol. (c) Raman spectra around the characteristic methanol C-O stretching wavelength in Cragganmore spiked with methanol.}
    \label{fig:PLSR_loading} 
\end{figure}

We tested machine learning regression models such as PCR, PLSR, and ridge regression to develop a model that could handle data from new whiskies with unknown methanol concentrations. PCR, PLSR, and ridge regression were used to create multivariate models to measure the levels of methanol. Machine learning regression models that were used in this study are described in the supplementary material section S4. The results are summarized in Table \ref{tbl:methanol_regression}. The best results were achieved with PLSR using mean-centering (PLSR7), which had $R^2_P$ = 0.997 and ${RMSE_P}$ = 0.05\% (in terms of methanol in the sample) (Figure \ref{fig:MeOH}). In particular, the PLSR7 model exhibited high performance for both the training and test sets, with the same $R^2$ (0.997) and RMSE (0.05\%) values. The model demonstrated its effective learning and generalization capabilities by performing well on a different set of whisky samples that were not included in the training set, indicating the model can handle various types of whisky samples. The peak close to 1020 \si{cm^{-1}} was assigned to the methanol C-O stretching vibration  and was the most significant feature contributing to this model, according to comparison the PLSR loading plot and the Raman spectra of methanol (Figure \ref{fig:PLSR_loading}). It should be noted that the methanol peak intensity could not be tracked directly due to its relative proximity to the ethanol peaks, as shown in Figure \ref{fig:EtOH_MeOH}.

Overall, the accurate quantification of methanol in whisky was made possible by PLSR and PCR with mean-centering, which also demonstrated acceptable expansion performance (see PLSR7 and PCR7 in Table \ref{tbl:methanol_regression}). Depending on the outcomes, polynomial baseline correction may be included, whereas vector normalization is not necessary to create methanol prediction models using PCR and PLSR. The PLSR7 model was able to detect methanol concentrations in whiskies that were as low as 0.17\%. This detection limit was based on the standard deviation of the response (\cite{guideline2005validation}). This detection limit for methanol is significantly lower than the maximum tolerable concentration of 2\% methanol in a spirit drink with 40\% alcohol by volume, which is considered safe for human consumption (\cite{RN497}). Additionally, the detection limit is even lower than the current EU general limit for naturally occurring methanol, which is 10 g methanol/L ethanol (equivalent to 0.4\% (v/v) methanol at 40\% alcohol) (\cite{RN497}).

\subsubsection{Through-bottle whisky brand identification}

To explore the broader applicability and robustness of these methods, the machine learning methods including KNN, LDA, ANN, RF, and RBF SVM were applied to a \emph{through-bottle} dataset. Each dataset was evaluated using each model, both with and without the inclusion of a 6 component PCA model to determine its predictive power. The best outcomes of the experiments are shown in Table \ref{tab:prelim_ttb}, with the full set of results in Table \ref{tab:Through-bottle}. In particular, the KNN model was able to achieve 100\% test accuracy with a dataset obtained only through bottles, as well as with a separate dataset that included measurements obtained through both vials and bottles. It should be noted that, for the instances where perfect prediction was achieved, many other models were also capable of producing comparable results.

The results obtained with this preliminary dataset provide valuable insights into the relative performance and predictability of the techniques used. Deep learning models were not applied in this case due to their need for larger volumes of data to produce meaningful results.

\begin{table}[H]
    \centering
    \caption{The best outcomes of through-bottle brand identification}
    \renewcommand{\arraystretch}{1.2}
    \begin{threeparttable}
    \begin{tabular}{lllll}
    \hline
        \textbf{Dataset\tnote{a}} & \textbf{Best Model} & \textbf{Accuracy} (\%) \\
    \hline
        VV  & KNN, RF, PCA+KNN, PCA+LDA, PCA+RF  & 100.0 \\
        TT  & KNN, PCA+KNN & 100.0 \\
        VT  & RF & 39.7 \\
        TV  & PCA + ANN & 52.3 \\
        Mix\tnote{b} & KNN, PCA+KNN, PCA+RF & 100.0 \\
    \hline
        \end{tabular}
        \begin{tablenotes}
      \item[a] The datasets are named using the format "$<$Train$><$Test$>$". Raman spectra were obtained either through vials (denoted as ‘V’) or through bottles (denoted as ‘T’). For example, if the model was trained using measurements obtained through vials and tested using measurements obtained through bottles, the dataset would be named ‘VT’
      \item[b] The training and test datasets include measurements obtained through both vials and bottles.
       \end{tablenotes}
        \end{threeparttable}
    \label{tab:prelim_ttb}
\end{table}

\section{Discussion}
Three deep learning methods were used to classify whisky brands and to quantify ethanol or methanol, and  nine conventional machine learning techniques were used for whisky brand identification, and three regression models for methanol quantification. The accuracy of these approaches was compared when the spectral data was either unprocessed or preprocessed using PCA. The deep learning models and three regression models for chemical regression were also compared. The results on the test set are summarized in Table \ref{tbl:ML_class}.

Several traditional machine learning algorithms such as KNN, RF, and LDA, as well as deep learning algorithms such as CNN and HPM, demonstrated excellent performance of $>$94\% using unprocessed data without PCA. It appears that the use of PCA improved the test accuracy of several conventional machine learning models, including QDA and ANN. Conventional machine learning models such as KNN, RF, and LDA, as well as deep learning models like CNN and HPM, achieved the best performance both with and without the use of PCA. The QDA and FCN models also performed well when PCA was used. The conventional machine learning models KNN and LDA performed very well without the use of PCA. However, while these conventional classification models are not be applicable for regression analysis, the deep learning models used in this study can also be applied to chemical regression. On the unseen ethanol test set, only the PCA+FCN model demonstrated good performance, achieving an $R^2$ value of 0.863 and an RMSE of 2.47\%. In contrast, none of the deep learning models exhibited good performance on the unseen methanol test set while the three conventional regression models excelled on the unseen methanol test set, with $R^2$ values greater than or equal to 0.995 and RMSEs greater than or equal to 0.07\%. 

\begin{table}[H]
\caption{Comparison table of brand identification and chemical regression on the test set.}
\renewcommand{\arraystretch}{1.2}
\begin{threeparttable}
\resizebox{\textwidth}{!}{%
\begin{tabular}{llllllll}
\hline
Model            & & without PCA & & & & with PCA\tnote{a} & \\  \cline{2-4} \cline{6-8}
                 & Accuracy (\%) & R\textsuperscript{2} (RMSE) & R\textsuperscript{2} (RMSE) & & Accuracy (\%) & R\textsuperscript{2} (RMSE) & R\textsuperscript{2} (RMSE) \\ 
                 & for brand & for ethanol & for methanol & & for brand & for ethanol & for methanol \\ \hline
CNN              & {\cellcolor{green_col}} 95.1        & {\cellcolor{red_col}} 0.342 (6.04)                            & {\cellcolor{red_col}} 0.092 (1.30)                             &  & {\cellcolor{green_col}} 96.6  & {\cellcolor{red_col}} 0.574 (4.71)                            & {\cellcolor{red_col}} 0.246 (1.13)                             \\
FCN              &  90.2        & {\cellcolor{red_col}} 0.032 (7.40)                            & {\cellcolor{red_col}} 0.007 (1.06)                             &  & {\cellcolor{green_col}} 95.1  & {\cellcolor{green_col}} 0.863 (2.47)                            & {\cellcolor{red_col}} 0.601 (1.00)                             \\
HPM              & {\cellcolor{green_col}} 94.8        & {\cellcolor{red_col}} 0.253 (7.06)                            &{\cellcolor{red_col}} 0.163 (1.30)                             &  & {\cellcolor{green_col}} 97.9  & {\cellcolor{red_col}} 0.645 (4.34)                            & {\cellcolor{red_col}} 0.300 (0.98)                             \\ \hline
Linear SVM       & {\cellcolor{red_col}} 20.7        & {\cellcolor{pink_col}} N/A                                    & {\cellcolor{pink_col}} N/A                                     &  & {\cellcolor{red_col}} 53.2  & {\cellcolor{pink_col}} N/A                                    & {\cellcolor{pink_col}} N/A                                     \\
RBF SVM          & {\cellcolor{red_col}} 8.1         & {\cellcolor{pink_col}} N/A                                    & {\cellcolor{pink_col}} N/A                                     &  & {\cellcolor{red_col}} 8.1 & {\cellcolor{pink_col}} N/A                                    & {\cellcolor{pink_col}} N/A                                     \\
KNN              & {\cellcolor{green_col}} 98.9        & {\cellcolor{pink_col}} N/A                                    & {\cellcolor{pink_col}} N/A                                     &  & {\cellcolor{green_col}} 98.6  & {\cellcolor{pink_col}} N/A                                    & {\cellcolor{pink_col}} N/A                                     \\
RF               & {\cellcolor{green_col}} 96.0        & {\cellcolor{pink_col}} N/A                                    & {\cellcolor{pink_col}} N/A                                     &  & {\cellcolor{green_col}} 99.7  & {\cellcolor{pink_col}} N/A                                    & {\cellcolor{pink_col}} N/A                                     \\
AdaBoost         & {\cellcolor{red_col}} 27.6        & {\cellcolor{pink_col}} N/A                                    & {\cellcolor{pink_col}} N/A                                     &  & {\cellcolor{red_col}} 14.4  & {\cellcolor{pink_col}} N/A                                    & {\cellcolor{pink_col}} N/A                                     \\
QDA              & {\cellcolor{red_col}} 24.7        & {\cellcolor{pink_col}} N/A                                    & {\cellcolor{pink_col}} N/A                                     &  & {\cellcolor{green_col}} 100.0   & {\cellcolor{pink_col}} N/A                                    & {\cellcolor{pink_col}} N/A                                     \\
LDA              & {\cellcolor{green_col}} 100.0       & {\cellcolor{pink_col}} N/A                                    & {\cellcolor{pink_col}} N/A                                     &  & {\cellcolor{green_col}} 100.0   & {\cellcolor{pink_col}} N/A                                    & {\cellcolor{pink_col}} N/A                                     \\
ANN              & {\cellcolor{red_col}} 3.5         & {\cellcolor{pink_col}} N/A                                    & {\cellcolor{pink_col}} N/A                                     &  &  84.5  & {\cellcolor{pink_col}} N/A                                    & {\cellcolor{pink_col}} N/A                                     \\ 
Gaussian Process & {\cellcolor{red_col}} 9.2         & {\cellcolor{pink_col}} N/A                                    & {\cellcolor{pink_col}} N/A                                     &  & {\cellcolor{red_col}} 9.2 & {\cellcolor{pink_col}} N/A                                    & {\cellcolor{pink_col}} N/A                                     \\ \hline
PCR              & {\cellcolor{pink_col}} N/A        & A                                       & {\cellcolor{green_col}} 0.997 (0.05)                             &  &                                            &                                         &                                          \\
PLSR             & {\cellcolor{pink_col}} N/A        & A                                       & {\cellcolor{green_col}} 0.997 (0.05)                             &  &                                            &                                         &                                          \\
Ridge regression & {\cellcolor{pink_col}} N/A        & A                                       & {\cellcolor{green_col}} 0.995 (0.07)                             &  &                                            &                                         &      \\ \hline                               
\end{tabular}%
}
    \begin{tablenotes}
      \item[a] Six PCA features were used as input for classification and regression
      \item N/A (pink) -  not applicable
      \item A - applicable but not tested
      \item Green - excellent performance
      \item Red - poor performance
    \end{tablenotes}
\end{threeparttable}
\label{tbl:ML_class}
\end{table}

The findings of this study can be used to develop new methods for the quantification of ethanol in whisky, and to improve the quality and efficiency of whisky production. In particular it is useful to note that the conventional machine learning and deep learning techniques are advantageous in different applications, particularly when there are varied quantities of traning data available. For example if the goal is to perform both brand identification and chemical regression simultaneously, a deep learning model may be the most suitable choice. However, if the task only requires either brand identification or chemical regression, then conventional machine learning models may result in equivalent or higher accuracy. It is also important to note that in order to achieve optimal performance with conventional machine learning models, it is necessary to carefully optimize them based on the results of spectral preprocessing and the number of PCA or PLS features used.

 Although the deep learning model has the ability to classify and predict many categories and contaminant levels simultaneously to a high precision, it takes vastly more data and computation to develop this knowledge base than traditional techniques. For example, when it comes to detecting methanol concentrations, a inherently challenging task given the low concentrations and the small changes to the spectra, this technique in its current form struggles compared to a more hands-on technique. Manual spectral windowing would likely improve performance, however this requires prior knowledge of the expected spectral changes which we have generally avoided with the goal of developing a fleixible analysis approach. As such, it is recommended that if this algorithm detects a low quantity of methanol, a more dedicated test be performed.

Machine learning methods have shown strong performance in identifying brands even when applied to datasets obtained through bottles, as summarized in Table \ref{tab:prelim_ttb}. These results suggest that when the training and testing sets are consistent, such as not being trained on through-bottle data and tested on through-vial data, the prediction problem can be largely solved using the techniques outlined in this work. Notably, even when applied to different types of vessels, some level of predictability was still observed. This suggests that the key features of the whisky spectra remain prominent when compared between beam and vessel types. These results highlight the robustness of the models and methods described here, although further training and a more diverse dataset may be required for practical applications as although the through-the-bottle method reduces the glass background it does not eliminate it completely. Future work will investigate methods to separately record the glass fluorescence and Raman spectra, such that this can be integrated within the data analysis process to fully separate the contents Raman from the container.

\section{Conclusions}

In this study, conventional machine learning and deep learning proved to be effective methods for a spectroscopic discrimination between twenty-eight whisky brands. Spectra were obtained with a small volume and, aside from decanting from the bottles into vials, no sample preparation was required. Despite previous publications in the literature addressing the application of Raman spectroscopy and chemometric methods for whisky analysis, the method presented in this work does not require a manual preprocessing step and performed extremely well in terms of brand identification and ethanol prediction. KNN and LDA achieved over 98\% test accuracy with or without PCA, while RF and QDA performed excellently with PCA, achieving over 99\% test accuracy. The HPM with PCA (PCA+HPM) was found to be the best performing deep learning algorithm for both brand identification (test accuracy of 98\%) and ethanol content prediction (test $R^2$ of 0.998 and RMSE of 0.25\%). The FCN with PCA (PCA+FCN) model accurately predicted the ethanol content in new samples, which were not used for training the model, with an $R^2$ of 0.863 and RMSE of 2.47\%. The PLSR model has been shown to be effective for predicting methanol contents (test $R^2$ of 0.997, RMSE of 0.05\%, and the detection limit of 0.17\%).

Several machine learning and deep learning models have shown good performance on raw data without the need for spectral preprocessing. This indicates that these models have the ability to adapt to new products/applications by working with data in its original form. The new HPM model is capable of both identifying brands and quantifying ethanol at the same time, whereas conventional machine learning methods can only perform classification and regression separately. While conventional machine learning methods were successful in identifying brands using this specific dataset, deep learning may be necessary for larger and more complex data sets as it can automatically learn and extract relevant features from the data without requiring manual feature engineering. Additionally, the ability of deep learning models in ethanol and methanol quantification are expected to improve with larger training datasets, as more data can help the model better learn and generalize from the data. Our machine learning methods have demonstrated their ability to be directly transferable to a through-bottle configuration, resulting in a system capable of obtaining spectra from unopened samples. The prediction accuracy of brand identification remained excellent even when applied to a small preliminary dataset of five whisky samples. This highlights the potential for practical applications of our methods in the analysis of unopened samples. 

For the primary analysis we chose a diverse set of twenty-eight whisky samples in terms of distilleries, flavours, cask types, ages, and ethanol contents in order to ensure that the training set is representative of real-world scenarios and collected data across different days to reduce environmental variability. Our chemical regression models for ethanol and methanol have demonstrated their ability to be applied to new samples. Specifically, the ethanol quantification model can predict the concentration of ethanol in gin samples and the methanol quantification model can effectively generalize to new or previously unseen samples.

Our technique has demonstrated its potential in detecting counterfeit spirits and assessing the quality of other high-value liquid samples. Our machine learning method was successful in accurately identifying whisky brands with subtle differences in ingredients, providing evidence of its effectiveness. The methods and techniques we developed should be applicable to detecting adulteration or substitution by training the model with a new dataset. Additionally, the lack of spectral preprocessing should enable this model to be applied directly to other high-value products such as edible oil, honey, or caviar. This represents a significant advancement in the field and opens up new possibilities for the detection of counterfeit and adulterated products.

\printbibliography

@misc{RN445,
    author = {{The British Broadcasting Corporation}},
   volume = {2023},
   number = {January},
   url = {https://www.bbc.com/news/uk-scotland-scotland-business-46566703},
   year = {2018},
   urldate = {2023-03-20},
   type = {Web Page}
}

@misc{RN446,
    author = {{The Times}},
   volume = {2023},
   number = {January},
   url = {https://www.thetimes.co.uk/article/dram-and-blast-third-of-vintage-scotch-whisky-found-to-befake-frhscnlx0},
   year = {2018},
   urldate = {2023-03-20},
   type = {Web Page}
}

@misc{RN447,
    author = {{The Guardian}},
   volume = {2023},
   number = {January},
   url = {https://www.theguardian.com/uk-news/2018/dec/20/rare-whisky-market-flooded-with-fakes-says-dealer},
   year = {2018},
   urldate = {2023-03-20},
   type = {Web Page}
}

@misc{Toxic2020,
    author = {{Tech Times}},
   volume = {2023},
   number = {January},
   url = {https://www.techtimes.com/articles/248428/20200329/social-media-misinformation-that-led-to-300-deaths-in-iran-claimed-drinking-methanol-was-a-cure-for-covid-19.htm},
   year = {2020},
   urldate = {2023-03-20},
   type = {Web Page}
}

@misc{Toxic2019,
    author = {{CNN}},
    volume = {2023},
    number = {January},
    url = {https://edition.cnn.com/2019/02/24/asia/india-alcohol-poisoning/index.html},
    year = {2019},
    urldate = {2023-03-20},
    type = {Web Page}
}

@techreport{FAP1,
   author = {{Fraud Advisory Panel}},
   title = {What’s your poison? The true cost of fake alcohol and how to catch the culprits},
   url = {https://www.fraudadvisorypanel.org/wp-content/uploads/2023/01/FAP-Special-Report-on-Alcohol-Fraud\_WEB.pdf},
   year = {2022},
   urldate = {2023-03-20},
   type = {Report}
}

@article{RN443,
   author = {Chaudhry, P. E. and Zimmerman, A. and Peters, J. R. and Cordell, V. V.},
   title = {Preserving intellectual property rights: Managerial insight into the escalating counterfeit market quandary},
   journal = {Business Horizons},
   volume = {52},
   number = {1},
   pages = {57-66},
   year = {2009},
   type = {Journal Article}
}

@article{RN453,
   author = {de Oliveira, L. P. and Rocha, D. P. and de Araujo, W. R. and Munoz, R. A. A. and Paixao, T. R. L. C. and Salles, M. O.},
   title = {Forensics in hand: new trends in forensic devices (2013-2017)},
   journal = {Analytical Methods},
   volume = {10},
   number = {43},
   pages = {5135-5163},
   year = {2018},
   type = {Journal Article}
}

@article{RN460,
   author = {Ellis, D. I. and Eccles, R. and Xu, Y. and Griffen, J. and Muhamadali, H. and Matousek, P. and Goodall, I. and Goodacre, R.},
   title = {Through-container, extremely low concentration detection of multiple chemical markers of counterfeit alcohol using a handheld SORS device},
   journal = {Scientific Reports},
   volume = {7},
   pages={12082},
   year = {2017},
   type = {Journal Article}
}

@article{RN459,
   author = {Ellis, D. I. and Muhamadali, H. and Xu, Y. and Eccles, R. and Goodall, I. and Goodacre, R.},
   title = {Rapid through-container detection of fake spirits and methanol quantification with handheld Raman spectroscopy},
   journal = {Analyst},
   volume = {144},
   number = {1},
   pages = {324-330},
   year = {2019},
   type = {Journal Article}
}

@article{RN444,
   author = {Green, R. T. and Smith, T.},
   title = {Executive insights: Countering brand counterfeiters},
   journal = {Journal of International Marketing},
   volume = {10},
   number = {4},
   pages = {89-106},
   year = {2002},
   type = {Journal Article}
}

@article{RN455,
   author = {Kiefer, J. and Cromwell, A. L.},
   title = {Analysis of single malt Scotch whisky using Raman spectroscopy},
   journal = {Analytical Methods},
   volume = {9},
   number = {3},
   pages = {511-518},
   year = {2017},
   type = {Journal Article}
}

@article{RN452,
   author = {Limm, W. and Karunathilaka, S. R. and Yakes, B. J. and Mossoba, M. M.},
   title = {A portable mid-infrared spectrometer and a non-targeted chemometric approach for the rapid screening of economically motivated adulteration of milk powder},
   journal = {International Dairy Journal},
   volume = {85},
   pages = {177-183},
   year = {2018},
   type = {Journal Article}
}

@article{RN442,
   author = {Power, A. C. and Ni Neill, C. and Geoghegan, S. and Currivan, S. and Deasy, M. and Cozzolino, D.},
   title = {A brief history of whiskey adulteration and the role of spectroscopy combined with chemometrics in the detection of modern whiskey fraud},
   journal = {Beverages},
   volume = {6},
   number = {3},
   pages = {49},
   year = {2020},
   type = {Journal Article}
}

@article{RN463,
   title = {Whisky analysis by Raman spectroscopy},
   journal = {Spectroscopy Europe},
   volume = {34},
   number = {5},
   pages = {40},
   year = {2022},
   type = {Journal Article}
}

@article{RN465,
   author = {Nordon, A. and Mills, A. and Burn, R. T. and Cusick, F. M. and Littlejohn, D.},
   title = {Comparison of non-invasive NIR and Raman spectrometries for determination of alcohol content of spirits},
   journal = {Analytica Chimica Acta},
   volume = {548},
   number = {1-2},
   pages = {148-158},
   year = {2005},
   type = {Journal Article}
}

@article{RN450,
   author = {Soon, J. M. and Manning, L.},
   title = {Developing anti-counterfeiting measures: The role of smart packaging},
   journal = {Food Research International},
   volume = {123},
   pages = {135-143},
   year = {2019},
   type = {Journal Article}
}

@inproceedings{germantrafficsigns2011,
  title={The German traffic sign recognition benchmark: a multi-class classification competition},
  author={Stallkamp, J. and Schlipsing, M. and Salmen, J. and Igel, C.},
  booktitle={The 2011 international joint conference on neural networks},
  pages={1453--1460},
  year={2011},
  organization={IEEE}
}

@article{krizhevsky_imagenet_2017,
	title = {{ImageNet} classification with deep convolutional neural networks},
	volume = {60},
	pages = {84--90},
	number = {6},
	journal = {Communications of the {ACM}},
	shortjournal = {Commun. {ACM}},
	author = {Krizhevsky, A. and Sutskever, I. and Hinton, G. E.},
    year = {2017},
}

@article{RN468,
   author = {Jimenez-Carvelo, A. M. and Osorio, M. T. and Koidis, A. and Gonzalez-Casado, A. and Cuadros-Rodriguez, L.},
   title = {Chemometric classification and quantification of olive oil in blends with any edible vegetable oils using FTIR-ATR and Raman spectroscopy},
   journal = {{LWT}-Food Science and Technology},
   volume = {86},
   pages = {174-184},
   year = {2017},
   type = {Journal Article}
}

@article{RN469,
   author = {Khan, S. and Ullah, R. and Khan, A. and Sohail, A. and Wahab, N. and Bilal, M. and Ahmed, M.},
   title = {Random Forest-based evaluation of Raman Spectroscopy for Dengue Fever analysis},
   journal = {Applied Spectroscopy},
   volume = {71},
   number = {9},
   pages = {2111-2117},
   year = {2017},
   type = {Journal Article}
}

@article{RN467,
   author = {Lussier, F. and Thibault, V. and Charron, B. and Wallace, G. Q. and Masson, J. F.},
   title = {Deep learning and artificial intelligence methods for Raman and surface-enhanced Raman scattering},
   journal = {{TrAC}-Trends in Analytical Chemistry},
   volume = {124},
   pages={115796},
   year = {2020},
   type = {Journal Article}
}

@article{RN475,
   author = {Kotsiantis, S. B.},
   title = {Supervised machine learning: A review of classification techniques},
   journal = {Informatica-Journal of Computing and Informatics},
   volume = {31},
   number = {3},
   pages = {249-268},
   year = {2007},
   type = {Journal Article}
}

@article{RN470,
   author = {Kotsiantis, S. B. and Zaharakis, I. D. and Pintelas, P. E.},
   title = {Machine learning: a review of classification and combining techniques},
   journal = {Artificial Intelligence Review},
   volume = {26},
   number = {3},
   pages = {159-190},
   year = {2006},
   type = {Journal Article}
}

@article{RN471,
   author = {Singh, A. and Thakur, N. and Sharma, A.},
   title = {A review of supervised machine learning algorithms},
   journal = {Proceedings of the 10th Indiacom - 2016 3rd International Conference on Computing for Sustainable Global Development},
   pages = {1310-1315},
   year = {2016},
   type = {Journal Article}
}

@article{RN484,
   author = {Liland, K. H. and Almoy, T. and Mevik, B. H.},
   title = {Optimal choice of baseline correction for multivariate calibration of spectra},
   journal = {Applied Spectroscopy},
   volume = {64},
   number = {9},
   pages = {1007-1016},
   year = {2010},
   type = {Journal Article}
}

@article{RN491,
   author = {Fan, X. Q. and Ming, W. and Zeng, H. T. and Zhang, Z. M. and Lu, H. M.},
   title = {Deep learning-based component identification for the Raman spectra of mixtures},
   journal = {Analyst},
   volume = {144},
   number = {5},
   pages = {1789-1798},
   year = {2019},
   type = {Journal Article}
}

@article{RN493,
   author = {LeCun, Y. and Bengio, Y. and Hinton, G.},
   title = {Deep learning},
   journal = {Nature},
   volume = {521},
   number = {7553},
   pages = {436-444},
   year = {2015},
   type = {Journal Article}
}

@article{RN492,
   author = {Pan, L. R. and Pipitsunthonsan, P. and Daengngam, C. and Channumsin, S. and Sreesawet, S. and Chongcheawchamnan, M.},
   title = {Identification of complex mixtures for Raman spectroscopy using a novel scheme based on a new multi-label deep neural network},
   journal = {IEEE Sensors Journal},
   volume = {21},
   number = {9},
   pages = {10834-10843},
   year = {2021},
   type = {Journal Article}
}

@article{RN497,
   author = {Paine, A. J. and Dayan, A. D.},
   title = {Defining a tolerable concentration of methanol in alcoholic drinks},
   journal = {Human \& Experimental Toxicology},
   volume = {20},
   number = {11},
   pages = {563-568},
   year = {2001},
   type = {Journal Article}
}

@article{RN498,
   author = {Burikov, S. and Dolenko, T. and Patsaeva, S. and Starokurov, Y. and Yuzhakov, V.},
   title = {Raman and IR spectroscopy research on hydrogen bonding in water-ethanol systems},
   journal = {Molecular Physics},
   volume = {108},
   number = {18},
   pages = {2427-2436},
   year = {2010},
   type = {Journal Article}
}

@article{RN499,
   author = {Lee, K. Y. M. and Paterson, A. and Piggott, J. R. and Richardson, G. D.},
   title = {Origins of flavour in whiskies and a revised flavour wheel: a review},
   journal = {Journal of the Institute of Brewing},
   volume = {107},
   number = {5},
   pages = {287-313},
   year = {2001},
   type = {Journal Article}
}

@article{RN519,
   author = {Cortes, C. and Vapnik, V.},
   title = {Support-Vector Networks},
   journal = {Machine Learning},
   volume = {20},
   number = {3},
   pages = {273-297},
   year = {1995},
   type = {Journal Article}
}

@article{RN520,
   author = {Ding, Y. and Zhu, H. Y. and Chen, R. Y. and Li, R. H.},
   title = {An efficient AdaBoost algorithm with the multiple thresholds classification},
   journal = {Applied Sciences},
   volume = {12},
   number = {12},
   pages = {5872},
   year = {2022},
   type = {Journal Article}
}

@article{RN518,
   author = {Dixon, S. J. and Brereton, R. G.},
   title = {Comparison of performance of five common classifiers represented as boundary methods: Euclidean Distance to Centroids, Linear Discriminant Analysis, Quadratic Discriminant Analysis, Learning Vector Quantization and Support Vector Machines, as dependent on data structure},
   journal = {Chemometrics and Intelligent Laboratory Systems},
   volume = {95},
   number = {1},
   pages = {1-17},
   year = {2009},
   type = {Journal Article}
}

@article{RN516,
   author = {Gou, J. P. and Ma, H. X. and Ou, W. H. and Zeng, S. N. and Rao, Y. B. and Yang, H. B.},
   title = {A generalized mean distance-based k-nearest neighbor classifier},
   journal = {Expert Systems with Applications},
   volume = {115},
   pages = {356-372},
   ISSN = {0957-4174},
   year = {2019},
   type = {Journal Article}
}

@book{RN571,
   author = {Kassambara, A.},
   title = {Machine learning essentials: Practical guide in R},
   publisher = {CreateSpace Independent Publishing Platform},
   ISBN = {1986406857},
   year = {2018},
   type = {Book}
}

@article{RN503,
   author = {Kohonen, T.},
   title = {An introduction to neural computing},
   journal = {Neural Networks},
   volume = {1},
   number = {1},
   pages = {3-16},
   year = {1988},
   type = {Journal Article}
}

@article{RN515,
   author = {Liaw, A. and Wiener, M.},
   title = {Classification and regression by randomForest},
   journal = {R News},
   volume = {2},
   number = {3},
   pages = {18--22},
   url = {http://CRAN.R-project.org/doc/Rnews/},
   year = {2002},
   urldate = {2023-03-20},
   type = {Journal Article}
}

@article{RN579,
   author = {McDonald, G. C.},
   title = {Ridge regression},
   journal = {Wiley Interdisciplinary Reviews-Computational Statistics},
   volume = {1},
   number = {1},
   pages={93--100},
   year = {2009},
   type = {Journal Article}
}

@book{RN569,
   author = {Rasmussen, C. E. and Williams, C. K. I.},
   title = {Gaussian processes for machine learning},
   publisher = {MIT Press},
   series = {Adaptive computation and machine learning},
   year = {2006},
   type = {Book}
}

@book{chollet2017deep,
  title={Deep learning with Python},
  author={Chollet, F.},
  year={2017},
  publisher={Simon and Schuster}
}

@inproceedings{wang2016edge,
  title={Edge detection using convolutional neural network},
  author={Wang, R.},
  booktitle={Advances in Neural Networks--ISNN 2016: 13th International Symposium on Neural Networks, ISNN 2016, St. Petersburg, Russia, July 6-8, 2016, Proceedings 13},
  pages={12--20},
  year={2016},
  organization={Springer}
}

@article{arbelaez2010contour,
  title={Contour detection and hierarchical image segmentation},
  author={Arbelaez, P. and Maire, M. and Fowlkes, C. and Malik, J.},
  journal={IEEE transactions on pattern analysis and machine intelligence},
  volume={33},
  number={5},
  pages={898--916},
  year={2010},
  publisher={IEEE}
}

@article{RN627,
   author = {Lunter, D. and Klang, V. and Kocsis, D. and Varga-Medveczky, Z. and Berko, S. and Erdo, F.},
   title = {Novel aspects of Raman spectroscopy in skin research},
   journal = {Experimental Dermatology},
   volume = {31},
   number = {9},
   pages = {1311-1329},
   year = {2022},
   type = {Journal Article}
}

@article{RN634,
   author = {Bzdok, D. and Altman, N. and Krzywinski, M.},
   title = {Statistics versus machine learning},
   journal = {Nature Methods},
   volume = {15},
   number = {4},
   pages = {233-234},
   year = {2018},
   type = {Journal Article}
}

@article{pca2014tutorial,
  title={A tutorial on principal component analysis},
  author={Shlens, J.},
  journal={arXiv preprint arXiv:1404.1100},
  year={2014}
}

@article{fleming2020through,
  title={Through-bottle whisky sensing and classification using Raman spectroscopy in an axicon-based backscattering configuration},
  author={Fleming, H. and Chen, M. and Bruce, G. D. and Dholakia, K.},
  journal={Analytical Methods},
  volume={12},
  number={37},
  pages={4572--4578},
  year={2020},
  publisher={Royal Society of Chemistry}
}

@article{shillito2022focus,
  title={To focus-match or not to focus-match inverse spatially offset Raman spectroscopy: a question of light penetration},
  author={Shillito, G. E. and McMillan, L. and Bruce, G. D. and Dholakia, K.},
  journal={Optics Express},
  volume={30},
  number={6},
  pages={8876--8888},
  year={2022},
  publisher={Optica Publishing Group}
}

@article{wang2020chemical,
  title={The chemical aspects of Raman spectroscopy: Statistical structure-spectrum relationship in the analyses of bioflavonoids},
  author={Wang, C.-H. and Huang, C.-C. and Chen, W. and Lai, Y.-S.},
  journal={Journal of Food and Drug Analysis},
  volume={28},
  number={2},
  pages={239},
  year={2020},
  publisher={Food and Drug Administration, Taiwan}
}

@article{lednev2012raman,
  title={Raman spectroscopy and advanced statistics for biochemical research and analytical purposes},
  author={Lednev, I. K. and Sikirzhytski, V.},
  journal={American Pharmaceutical Review},
  volume={15},
  number={4},
  year={2012},
  publisher={Russell Publishing LLC}
}

@incollection{singh2021diagnosing,
  title={Diagnosing of disease using machine learning},
  author={Singh, P. and Singh, N. and Singh, K. K. and Singh, A.},
  booktitle={Machine learning and the internet of medical things in healthcare},
  pages={89--111},
  year={2021},
  publisher={Elsevier}
}

@article{guideline2005validation,
  title={Validation of analytical procedures: text and methodology},
  author={{ICH Harmonised Tripartite Guideline}},
  journal={Q2 (R1)},
  volume={1},
  number={20},
  pages={05},
  year={2005}
}

@article{sarker2021deep,
  title={Deep learning: a comprehensive overview on techniques, taxonomy, applications and research directions},
  author={Sarker, I. H.},
  journal={SN Computer Science},
  volume={2},
  number={6},
  pages={420},
  year={2021},
  publisher={Springer}
}

@inproceedings{pal2020performance,
  title={Performance Evaluation of Multivariate Linear Regression Model with LDA and PCA},
  author={Pal, K. and Sharma, M.},
  booktitle={Proceedings of the International Conference on Recent Advances in Computational Techniques (IC-RACT)},
  year={2020}
}

@article{janiesch2021machine,
  title={Machine learning and deep learning},
  author={Janiesch, C. and Zschech, P. and Heinrich, K.},
  journal={Electronic Markets},
  volume={31},
  number={3},
  pages={685--695},
  year={2021},
  publisher={Springer}
}

@article{ashok2013optofluidic,
  title={Optofluidic Raman sensor for simultaneous detection of the toxicity and quality of alcoholic beverages},
  author={Ashok, Praveen C and Praveen, Bavishna B and Dholakia, Kishan},
  journal={Journal of Raman Spectroscopy},
  volume={44},
  number={6},
  pages={795--797},
  year={2013},
  publisher={Wiley Online Library}
}

@article{ashok2011near,
  title={Near infrared spectroscopic analysis of single malt Scotch whisky on an optofluidic chip},
  author={Ashok, Praveen C and Praveen, Bavishna B and Dholakia, K},
  journal={Optics express},
  volume={19},
  number={23},
  pages={22982--22992},
  year={2011},
  publisher={Optica Publishing Group}
}
\newpage
\beginsupplement
\section*{Supplementary Material}
\subsection{Spectral characteristics}
\begin{figure}[ht]
    \begin{subfigure}{0.57\textwidth}
        \caption{}
        \centering
        \includegraphics[width=1\textwidth]{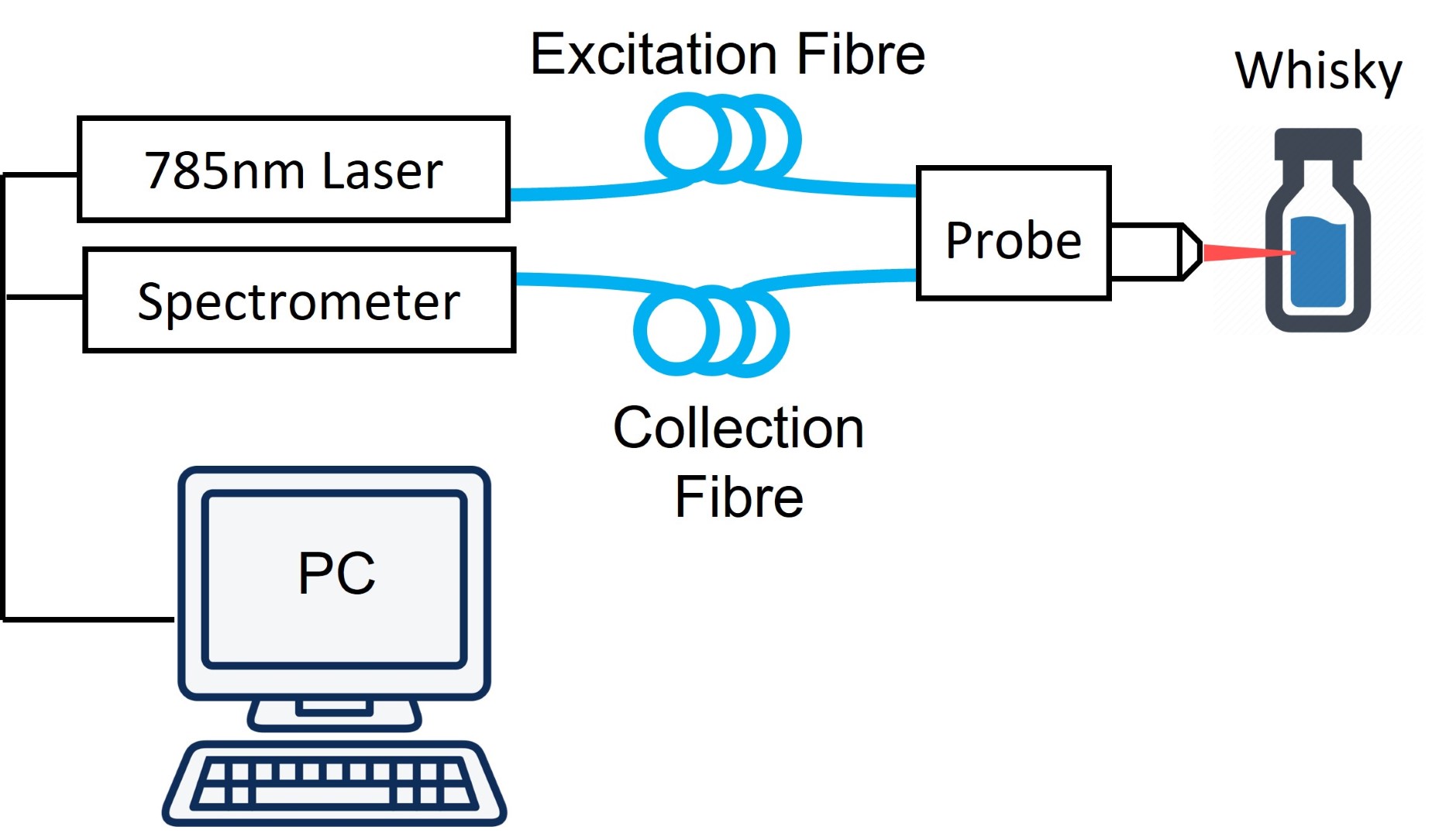} 
        \label{fig:config}
    \end{subfigure}\hfill
    \begin{subfigure}{0.43\textwidth}
        \caption{}
        \centering
        \includegraphics[width=1\textwidth]{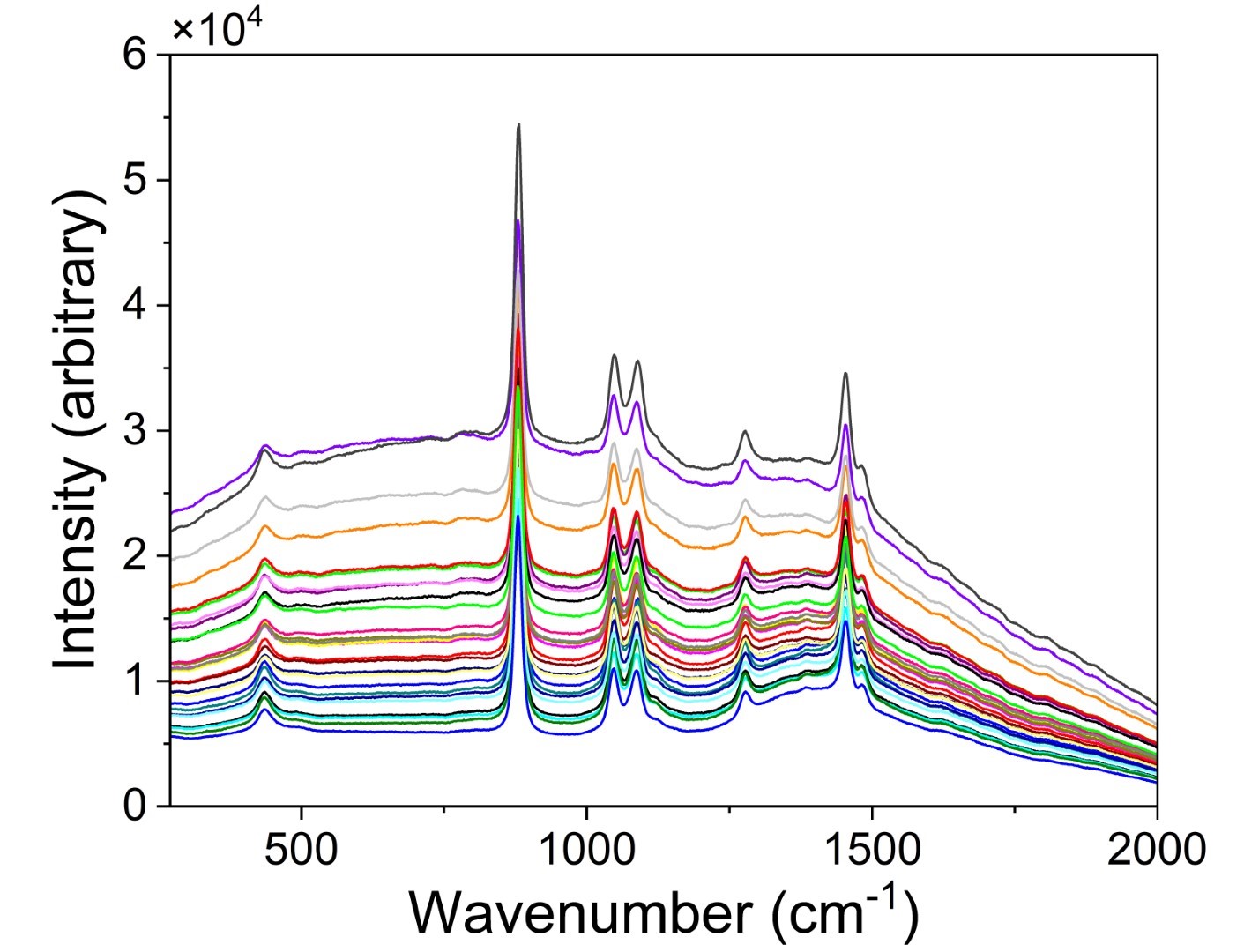} 
        \label{fig:Spectra}
    \end{subfigure}
    \caption{(a) Experimental configuration. A 785 nm laser was coupled into the head of a commerical probe. This probe head contains filters and other optics required for Raman spectroscopy. The sample vial is placed in front of the probe with the signal collected by the spectrometer. (b) Representative Raman spectra of 28 commercial whisky samples, demonstrating sharp Raman peaks corresponding to the ethanol content, and variable fluorescence signature from natural/artificial colours in the whisky. }
    \label{fig:Raman}
\end{figure}

The spectra of samples were measured using a portable Raman spectrometer with the schematic shown in Figure \ref{fig:config}. Figure \ref{fig:Spectra} displays the representative Raman spectra of 28 samples, demonstrating the different spectral signatures that can be found in the samples. The sharp features are primarily due to the vibrational modes of ethanol 
(\cite{RN498}). Whiskies also contain phenolic compounds, aldehydes, esters, and other chemicals known as congeners, which are produced during the fermentation or distillation process (\cite{RN499}). These are responsible for the diversity and complexity of flavours, aromas, and colours in whiskies. The Raman signals appear on top of a broad fluorescence background (Figure \ref{fig:Spectra}). Because the fluorescence background is considered to be caused by congeners, this fluorescence signal can provide additional information about the sample. 

\subsection{Deep Learning Methods}
The network has two main branches, a convolutional (CNN) and a fully connected (FCN) branch. The networks were trained and evaluated at multiple stages of training to obtain contextual understanding of their relative robustness and effectiveness. CNN has been shown to be highly effective in image classification and has won several high profile competitions in this field (\cite{germantrafficsigns2011, krizhevsky_imagenet_2017}). The convolutional layers in CNN analyze small portions of an image, and these local features are combined to form more complex and sophisticated representations. This approach is well-suited for identifying peaks within a spectrum, but tends to overlook the broader fluorescence background. In contrast, the series of fully connected network (FCN) excel at capturing the fluorescence background, but struggle to identify finer features such as peaks (\cite{arbelaez2010contour, wang2016edge}). These layers are also computationally intensive and tend to become too flexible before they have sufficient predictive power, due to the large solution space that the back-propagation algorithm must search as the network is not specifically designed for a particular task. The proposed combination of the two, hybrid parallel model (HPM), is expected to allow for the information from each side to be distilled and propagated through, allowing for a more complete representation of the data. As this objective here is to mitigate information loss, it is vital that both these blocks of layers have access to the spectra input directly. After feeding through these submodels, the results are concatenated back as to allow the attached classifiers to have access to the full richness of the combined data abstractions.

\subsection{Conventional Machine Learning Methods for brand identification}
SVM is a type of binary linear classification that uses a non-linear step referred to as the kernel transformation (\cite{RN519}). The input spectral space is transformed into a feature space through a mathematical transformation and a linear decision boundary is established between samples near the border of each class. Two kernel functions, linear and radial basis function (RBF) were applied to the SVM model. SVM with linear kernel and SVM with RBF kernel are abbreviated as linear SVM and RBF SVM, respectively. KNN is a widely used supervised learning method that determines categories by finding the maximum number of categories within a specific range determined by the Euclidean distance (\cite{RN516}). RF is a non-linear ensemble method consisting of multiple decision trees classified through voting based on the Gini index (\cite{RN515}). Adaptive Boost (AdaBoost) is a popular ensemble learning algorithm that combines weak classifiers into strong classifiers through weighted majority voting (\cite{RN520}). Quadratic Discriminant Analysis (QDA) is a discriminant analysis algorithm based on Mahalanobis distance calculations between samples for each class (\cite{RN518}). This algorithm can also take into account class size differences using Bayesian probability terms. Linear Discriminant Analysis (LDA) can be understood as a supervised algorithm that calculates linear combinations of features that represent the axes that best separate multiple class labels (\cite{pal2020performance}). Artificial Neural Network (ANN) is a massively parallel interconnected network of simple elements and hierarchical organizations that function similarly to biological nervous systems (\cite{RN503}). Gaussian Process is a generalization of the Gaussian probability distribution that governs the properties of functions, whereas a probability distribution only describes random variables as scalars or vectors (\cite{RN569}).

\subsection{Machine learning for methanol quantification}
PCR involves reducing the original predictor variables into a smaller number of principal components through PCA and using these components to build the linear regression model (\cite{RN571}). PLSR, on the other hand, considers the relationship between the new principal components and the outcome to determine the components used in the regression model (\cite{RN571}). Unlike PCR, PLSR takes a dimension reduction approach that is guided by the outcome. Ridge regression, on the other hand, provides a solution to the issue of collinearity without eliminating any variables from the original set of independent variables (\cite{RN579}).

\begin{table}[ht]
  \caption{List of samples used for through the vial analysis}
  \label{tbl:whisky_samples}
  \resizebox{\textwidth}{!}{
  \renewcommand{\arraystretch}{1.2}
    \begin{tabular}{lllllllll}
        \hline
        Number  & Samples                           & Alcohol (\%) & Age (year) & Origin    & Type         \\
        \hline
        1       & Ardbeg, Wee Beastie               & 47.4         & 5          & Scotland  & Single malt  \\
        2       & Black Bull                        & 50           & 12         & Scotland  & Blended      \\
        3       & Bowmore - Islay                   & 40           & 12         & Scotland  & Single malt  \\
        4       & Cambus 1991, Signatory            & 55.3         & 26         & Scotland  & Single grain \\
        5       & Caol Ila                          & 43           & 12         & Scotland  & Single malt  \\
        6       & Clynelish                         & 46           & 14         & Scotland  & Single malt  \\
        7       & Cragganmore                       & 40           & 12         & Scotland  & Single malt  \\
        8       & Cragganmore, Special Release 2020 & 55.8         & 20         & Scotland  & Single malt  \\
        9       & Dalwhinnie                        & 43           & 15         & Scotland  & Single malt  \\
        10      & Glenfiddich “Distillery Edition”  & 51           & 15         & Scotland  & Single malt  \\
        11      & Glenkinchie                       & 43           & 12         & Scotland  & Single malt  \\
        12      & Glenlivet, 'Perth' Single Cask    & 51.3         & 18         & Scotland  & Single malt  \\
        13      & Glenlivet, 'Tom an Uird' Single Cask  & 58.4         & 16         & Scotland  & Single malt  \\
        14      & Ichiros - Malt \& Grain           & 46.5         &            & Japan     & Blended      \\
        15      & Kavalan - Conductor               & 46           &            & Taiwan    & Single malt  \\
        16      & Lagavulin                         & 43           & 16         & Scotland  & Single malt  \\
        17      & Mackmyra, Moment, Fjallmark       & 42           &            & Sweden    & Single malt  \\
        18      & Nant - Sherry Cask                & 43           &            & Australia & Single malt  \\
        19      & Nant, Port Wood                   & 63           &            & Australia & Single malt  \\
        20      & Nikka - Coffey Grain              & 45           &            & Japan     & Grain        \\
        21      & Nikka - Miyagikyo                 & 45           &            & Japan     & Single malt  \\
        22      & Nikka - Taketsuru                 & 43           &            & Japan     & Single malt  \\
        23      & Oban                              & 43           & 14         & Scotland  & Single malt  \\
        24      & Overeem, Sherry Cask Matured      & 43           &            & Australia & Single malt  \\
        25      & Overeem, Sherry Cask Strength     & 60           &            & Australia & Single malt  \\
        26      & Talisker                          & 45.8         & 10         & Scotland  & Single malt  \\
        27      & Tomintoul - Oloroso Sherry Cask   & 40           & 12         & Scotland  & Single malt  \\
        28      & Tomintoul - With A Peaty Tang     & 40           &            & Scotland  & Single malt  \\
        29      & 40\% Ethanol/Water                & 40           &            &           &              \\
        \hline
    \end{tabular}}
\end{table}

\begin{table}[ht]
  \caption{List of samples used for through the bottle analysis}
  \label{tbl:whisky_samples_ttb}
  \resizebox{\textwidth}{!}{
  \renewcommand{\arraystretch}{1.2}
    \begin{tabular}{lllllllll}
        \hline
        Number  & Samples                           & Alcohol (\%) & Age (year) & Origin    & Type         \\
        \hline
        1       & Dalwhinnie                        & 43           & 15         & Scotland  & Single malt  \\
        2       & Glenkinchie                       & 43           & 12         & Scotland  & Single malt  \\
        3       & Glengoyne                         & 43           & 12         & Scotland  & Single malt  \\
        4       & Glengoyne "The Legacy Chapter 2"  & 48           &            & Scotland  & Single malt  \\
        5       & Glengoyne                         & 43           & 18         & Scotland  & Single malt  \\     
        6       & Oban                              & 43           & 14         & Scotland  & Single malt  \\
        7       & The Famous Grouse                 & 40           &            & Scotland  & Blended  \\
        \hline
    \end{tabular}}
\end{table}

\begin{table}[ht]
  \caption{Splitting initial data into subsets for methanol quantification}
  \label{tbl:Methanol dataset}
  \resizebox{\textwidth}{!}{
    \begin{tabular*}{\textwidth}{p{4cm}p{9cm}}
    \hline
    Training and validation sets    & Talisker, Cragganmore, and 40\% ethanol/water spiked with methanol ( 0-3\% in 0.3\% increments) \\
    \hline
    Test set          &  Caol Ila and Cynelish spiked with methanol (0, 0.3, 1, and 2\%)  \\
    \hline
    \end{tabular*}}
\end{table}

\begin{table}[ht]
  \caption{Deep learning model hyperparameters for each block}
  \label{tbl:DL_parameters}
  \resizebox{\textwidth}{!}{
    \begin{tabular*}{\textwidth}{p{5cm}p{8cm}}
    \hline
    Model               & Parameters \\
    \hline
    FCN path            &  10 neurons, 2 layers \\
    CNN path            & 5 blocks, 2 conv. layers each + a 1D max pooling  \\
    CNN path (with PCA) & 5 blocks, 2 conv. layers each  \\
    Feature abstraction & 40 neurons, 2 layers  \\
    Brand classifier    & 30 neurons, 2 layers  \\
    Ethanol regression  & 10 neurons, 2 layers  \\
    Methanol regression & 10 neurons, 2 layers  \\
    Batches             & 32  \\
    \hline
    \end{tabular*}}
\end{table}

\begin{table}[ht]
  \caption{Conventional machine learning model parameters}
  \label{tbl:conv_ML_parameters}
  \resizebox{\textwidth}{!}{
    \begin{tabular*}{\textwidth}{p{4cm}p{9cm}}
    \hline
    Model               & Parameters \\
    \hline
    Linear SVM          & loss: \texttt{hinge}, intercept scaling: 1000  \\
    RBF SVM             & gamma: \texttt{auto}  \\
    KNN                 & Nearest neighbours: 1  \\
    RF                  & Maximum depth: \texttt{None}, Estimators: 100  \\
    AdaBoost            & None  \\
    QDA                 & None  \\
    LDA                 & None  \\
    ANN                 & Alpha: 1, Maximum iterations: 1000  \\
    Gaussian Process    & None  \\
    \hline
    \end{tabular*}}
\end{table}

\begin{table}[ht]
  \caption{Brand identification accuracy using deep leaning models}
  \begin{threeparttable}
  \resizebox{\textwidth}{!}{
    \begin{tabular}{llllll}
		\hline
	 &      & \multicolumn{3}{c}{Accuracy (\%)} &                                \\ \cline{3-5}
    Methods & No. of epochs & Training  & Validation  & Test    & Fit Time (s) \\
		\hline
        CNN	& 100	& 82.29	& 79.53	& 78.15	& 48 \\
        CNN	& 200	& 94.17	& 91.47	& 88.31	& 93 \\
        CNN	& 500	& 98.27	& 96.76	& 93.85	& 223 \\
        CNN	& 1000	& 99.78	& 98.47	& 95.08	& 439 \\
        CNN	& 2000	& 99.35	& 98.18	& 94.15	& 872 \\
        FCN	& 100	& 62.85	& 60.94	& 61.85	& 39 \\
        FCN	& 200	& 68.25	& 67.71	& 66.77	& 76 \\
        FCN	& 500	& 80.35	& 76.24	& 76.00	& 188 \\
        FCN	& 1000	& 92.01	& 90.00	& 86.77	& 371 \\
        FCN	& 2000	& 96.98	& 94.24	& 90.15	& 739 \\
        HPM	& 100	& 85.10	& 81.53	& 80.62	& 49 \\
        HPM	& 200	& 66.09	& 62.76	& 63.08	& 95 \\
        HPM	& 500	& 100.00	& 98.00	& 93.85	& 228 \\
        HPM	& 1000	& 99.78	& 99.24	& 94.77	& 452 \\
        HPM	& 2000	& 100.00	& 99.24	& 94.77	& 896 \\
        \hline
        PCA+CNN\tnote{a}	& 100	& 96.76	& 93.47	& 92.00	& 45 \\
        PCA+CNN\tnote{a}	& 200	& 100.00	& 98.41	& 96.31	& 87 \\
        PCA+CNN\tnote{a}	& 500	& 100.00	& 98.65	& 94.46	& 210 \\
        PCA+CNN\tnote{a}	& 1000	& 100.00	& 98.94	& 96.62	& 416 \\
        PCA+CNN\tnote{a}	& 2000	& 100.00	& 98.59	& 95.38	& 828 \\
        PCA+FCN\tnote{a}	& 100	& 87.69	& 84.18	& 85.85	& 37 \\
        PCA+FCN\tnote{a}	& 200	& 94.82	& 93.35	& 93.85	& 75 \\
        PCA+FCN\tnote{a}	& 500	& 98.92	& 96.88	& 95.38	& 182 \\
        PCA+FCN\tnote{a}	& 1000	& 100.00	& 97.88	& 94.77	& 360 \\
        PCA+FCN\tnote{a}	& 2000	& 100.00	& 97.88	& 95.08	& 718 \\
        PCA+HPM\tnote{a}	& 100	& 98.49	& 96.53	& 95.69	& 46 \\
        PCA+HPM\tnote{a}	& 200	& 98.92	& 97.71	& 97.23	& 89 \\
        PCA+HPM\tnote{a}	& 500	& 100.00	& 98.47	& 97.23	& 216 \\
        PCA+HPM\tnote{a}	& 1000	& 100.00	& 99.24	& 96.92	& 429 \\
        PCA+HPM\tnote{a}	& 2000	& 100.00	& 99.35	& 97.85	& 854 \\
		\hline
    \end{tabular}}
      \begin{tablenotes}
        \item[a] PCA was applied to the data before training the model.
    \end{tablenotes}
\end{threeparttable}
\label{tbl:deeplearning_class}
\end{table}

\begin{table}[ht]
  \caption{Ethanol quantification of deep leaning models}
  \begin{threeparttable}
  \resizebox{\textwidth}{!}{
  \renewcommand{\arraystretch}{1.2}
    \begin{tabular}{lllllllllll}
		\hline
	Methods      & No. of epochs  & \textbf{$R^2_T$}    & \textbf{${RMSE_T}$}   & \textbf{$R^2_V$}    & \textbf{${RMSE_V}$}   & \textbf{$R^2_P$}    & \textbf{${RMSE_P}$}   & \textbf{$R^2_U$}\tnote{b}    & \textbf{${RMSE_U}$}\tnote{b}  \\
		\hline
        CNN	& 100	& 0.8416	& 2.1965	& 0.8476	& 2.1036	& 0.8166	& 2.2670	& 0.0076	& 8.8172 \\
        CNN	& 200	& 0.9687	& 0.9723	& 0.9650	& 0.9969	& 0.9655	& 0.9599	& 0.0109	& 10.1917 \\
        CNN	& 500	& 0.9859	& 0.6419	& 0.9813	& 0.7199	& 0.9873	& 0.5816	& 0.3418	& 6.0397 \\
        CNN	& 1000	& 0.9972	& 0.2846	& 0.9947	& 0.3894	& 0.9911	& 0.4912	& 0.1764	& 7.2112 \\
        CNN	& 2000	& 0.9985	& 0.2074	& 0.9971	& 0.2893	& 0.9943	& 0.3917	& 0.2335	& 7.0251 \\
        FCN	& 100	& 0.3888	& 4.5173	& 0.3940	& 4.3365	& 0.3796	& 4.3068	& 0.0318	& 7.3979 \\
        FCN	& 200	& 0.6527	& 3.2720	& 0.6136	& 3.2956	& 0.6003	& 3.2966	& 0.0300	& 7.9259 \\
        FCN	& 500	& 0.9024	& 2.1156	& 0.8923	& 2.1435	& 0.8775	& 2.1253	& 0.0290	& 7.9859 \\
        FCN	& 1000	& 0.9600	& 1.2264	& 0.9520	& 1.2801	& 0.9442	& 1.2807	& 0.0012	& 8.8209 \\
        FCN	& 2000	& 0.9812	& 0.7686	& 0.9783	& 0.8000	& 0.9754	& 0.8204	& 0.0054	& 9.2550 \\
        HPM	& 100	& 0.9526	& 1.1900	& 0.9456	& 1.2547	& 0.9374	& 1.3281	& 0.0162	& 7.8303 \\
        HPM	& 200	& 0.8828	& 2.2671	& 0.8581	& 2.4587	& 0.8516	& 2.4474	& 0.2294	& 6.3492 \\
        HPM	& 500	& 0.9951	& 0.3864	& 0.9936	& 0.4292	& 0.9854	& 0.6306	& 0.2033	& 6.7653 \\
        HPM	& 1000	& 0.9987	& 0.1967	& 0.9981	& 0.2316	& 0.9929	& 0.4338	& 0.2526	& 7.0635 \\
        HPM	& 2000	& 0.9996	& 0.1036	& 0.9983	& 0.2206	& 0.9909	& 0.4932	& 0.2017	& 6.7536 \\ \hline
        PCA+CNN\tnote{a}	& 100	& 0.9859	& 0.6710	& 0.9817	& 0.7353	& 0.9771	& 0.8106	& 0.4965	& 5.1190 \\
        PCA+CNN\tnote{a}	& 200	& 0.9988	& 0.2337	& 0.9973	& 0.3068	& 0.9948	& 0.4172	& 0.5744	& 4.7132 \\
        PCA+CNN\tnote{a}	& 500	& 0.9995	& 0.1470	& 0.9983	& 0.2381	& 0.9955	& 0.3644	& 0.4605	& 5.8649 \\
        PCA+CNN\tnote{a}	& 1000	& 0.9999	& 0.0645	& 0.9986	& 0.1950	& 0.9978	& 0.2421	& 0.2204	& 7.5093 \\
        PCA+CNN\tnote{a}	& 2000	& 1.0000	& 0.0374	& 0.9989	& 0.1759	& 0.9964	& 0.3076	& 0.1952	& 8.5766 \\
        PCA+FCN\tnote{a}	& 100	& 0.9724	& 0.9635	& 0.9671	& 1.0099	& 0.9712	& 0.8916	& 0.8611	& 2.6153 \\
        PCA+FCN\tnote{a}	& 200	& 0.9784	& 0.8314	& 0.9743	& 0.8722	& 0.9790	& 0.7552	& 0.8634	& 2.4702 \\
        PCA+FCN\tnote{a}	& 500	& 0.9840	& 0.8364	& 0.9808	& 0.8466	& 0.9813	& 0.7584	& 0.5477	& 5.3886 \\
        PCA+FCN\tnote{a}	& 1000	& 0.9895	& 0.6568	& 0.9839	& 0.7396	& 0.9834	& 0.6911	& 0.5193	& 5.5237 \\
        PCA+FCN\tnote{a}	& 2000	& 0.9934	& 0.4515	& 0.9884	& 0.5755	& 0.9869	& 0.6223	& 0.3140	& 6.7286 \\
        PCA+HPM\tnote{a}	& 100	& 0.9845	& 0.7658	& 0.9827	& 0.7750	& 0.9808	& 0.8033	& 0.5089	& 4.8056 \\
        PCA+HPM\tnote{a}	& 200	& 0.9939	& 0.5442	& 0.9921	& 0.5795	& 0.9924	& 0.5676	& 0.5289	& 5.1535 \\
        PCA+HPM\tnote{a}	& 500	& 0.9996	& 0.1324	& 0.9981	& 0.2418	& 0.9975	& 0.2743	& 0.5816	& 4.5335 \\
        PCA+HPM\tnote{a}	& 1000	& 0.9998	& 0.0991	& 0.9987	& 0.2022	& 0.9970	& 0.2906	& 0.5521	& 5.0017 \\
        PCA+HPM\tnote{a}	& 2000	& 1.0000	& 0.0339	& 0.9992	& 0.1501	& 0.9976	& 0.2526	& 0.6454	& 4.3390 \\
		\hline
    \end{tabular}
    }
      \begin{tablenotes}
        \item[a] PCA was applied to the data before training the model.
        \item[b] Entirely new and unseen samples (e.g. different brands).
      \end{tablenotes}
\end{threeparttable}
\label{tbl:deeplearning_alcohol}
\end{table}

\begin{table}[ht]
  \caption{Methanol quantification using deep leaning models}
  \begin{threeparttable}
  \resizebox{\textwidth}{!}{
  \renewcommand{\arraystretch}{1.2}
    \begin{tabular}{llllllll}
		\hline
	Methods      & No. of epochs  & \textbf{$R^2_T$}    & \textbf{${RMSE_T}$}   & \textbf{$R^2_V$}    & \textbf{${RMSE_V}$}   & \textbf{$R^2_P$}    & \textbf{${RMSE_P}$}  \\
		\hline
		CNN	& 100	& 0.4886	& 0.7295	& 0.4536	& 0.7737	& 0.0692	& 1.2768 \\
		CNN	& 200	& 0.5558	& 0.6798	& 0.5099	& 0.7332	& 0.0897	& 1.3039 \\
		CNN	& 500	& 0.8888	& 0.3475	& 0.8669	& 0.3893	& 0.0921	& 1.3030 \\
		CNN	& 1000	& 0.9556	& 0.2488	& 0.9520	& 0.2501	& 0.0015	& 1.3031 \\
		CNN	& 2000	& 0.9948	& 0.0753	& 0.9918	& 0.0966	& 0.0522	& 1.3043 \\
		FCN	& 100	& 0.1242	& 0.9625	& 0.0844	& 1.0060	& 0.0039	& 0.7307 \\
		FCN	& 200	& 0.2176	& 0.9226	& 0.1682	& 0.9649	& 0.0036	& 0.8267 \\
		FCN	& 500	& 0.4094	& 0.7936	& 0.3773	& 0.8333	& 0.0067	& 1.0636 \\
		FCN	& 1000	& 0.5217	& 0.7041	& 0.4922	& 0.7456	& 0.0048	& 1.2950 \\
		FCN	& 2000	& 0.6004	& 0.6440	& 0.5794	& 0.6790	& 0.0050	& 1.3035 \\
		HPM	& 100	& 0.5083	& 0.7162	& 0.4738	& 0.7594	& 0.0659	& 1.2992 \\
		HPM	& 200	& 0.6842	& 0.5727	& 0.6238	& 0.6412	& 0.0040	& 1.3045 \\
		HPM	& 500	& 0.7417	& 0.5271	& 0.7017	& 0.5783	& 0.0618	& 1.3016 \\
		HPM	& 1000	& 0.8337	& 0.4147	& 0.7793	& 0.4906	& 0.1627	& 1.3043 \\
		HPM	& 2000	& 0.9805	& 0.1451	& 0.9771	& 0.1628	& 0.0071	& 1.3046 \\ \hline
		PCA+CNN\tnote{a}	& 100	& 0.6506	& 0.6367	& 0.6267	& 0.6691	& 0.0572	& 1.3041 \\
		PCA+CNN\tnote{a}	& 200	& 0.9143	& 0.2992	& 0.9031	& 0.3263	& 0.1715	& 1.3009 \\
		PCA+CNN\tnote{a}	& 500	& 0.9967	& 0.0601	& 0.9938	& 0.0833	& 0.1575	& 1.2997 \\
		PCA+CNN\tnote{a}	& 1000	& 0.9968	& 0.0698	& 0.9935	& 0.0932	& 0.2462	& 1.1272 \\
		PCA+CNN\tnote{a}	& 2000	& 0.9993	& 0.0276	& 0.9945	& 0.0800	& 0.1227	& 1.3016 \\
		PCA+FCN\tnote{a}	& 100	& 0.5508	& 0.6947	& 0.5142	& 0.7403	& 0.0571	& 1.2666 \\
		PCA+FCN\tnote{a}	& 200	& 0.8732	& 0.3708	& 0.8236	& 0.4457	& 0.2804	& 1.3042 \\
		PCA+FCN\tnote{a}	& 500	& 0.9405	& 0.2749	& 0.9173	& 0.3258	& 0.3687	& 1.3038 \\
		PCA+FCN\tnote{a}	& 1000	& 0.9883	& 0.1133	& 0.9826	& 0.1419	& 0.6684	& 1.2482 \\
		PCA+FCN\tnote{a}	& 2000	& 0.9799	& 0.1489	& 0.9809	& 0.1518	& 0.6011	& 0.9984 \\
		PCA+HPM\tnote{a}	& 100	& 0.5943	& 0.7476	& 0.6053	& 0.7345	& 0.0685	& 1.2910 \\
		PCA+HPM\tnote{a}	& 200	& 0.9848	& 0.1563	& 0.9844	& 0.1687	& 0.2986	& 1.1585 \\
		PCA+HPM\tnote{a}	& 500	& 0.9893	& 0.1295	& 0.9880	& 0.1428	& 0.2856	& 1.0800 \\
		PCA+HPM\tnote{a}	& 1000	& 0.9989	& 0.0356	& 0.9961	& 0.0656	& 0.2999	& 0.9820 \\
		PCA+HPM\tnote{a}	& 2000	& 0.9998	& 0.0158	& 0.9963	& 0.0650	& 0.2557	& 1.0600 \\

		\hline
    \end{tabular}
    }
      \begin{tablenotes}
        \item[a] PCA was applied to the data before training the model.
        \item
      \end{tablenotes}
\end{threeparttable}
\label{tbl:deeplearning_methanol}
\end{table}

\begin{table}[ht]
  \caption{Methanol quantification using different methods}
  \begin{threeparttable}
  \resizebox{\textwidth}{!}{
  \renewcommand{\arraystretch}{1.2}
    \begin{tabular}{llllllllll}
    \hline
\textbf{Model}   & \textbf{Method} & \textbf{\begin{tabular}[c]{@{}l@{}}Preprocessing\\ or parameter\end{tabular}}  & \textbf{Factors}\tnote{c}   & \textbf{$R^2_T$}    & \textbf{${RMSE_T}$}   & \textbf{$R^2_C$}    & \textbf{${RMSE_C}$}   & \textbf{$R^2_P$}    & \textbf{${RMSE_P}$}  \\
 \hline
PCR1   & PCR     & poly09\tnote{a}    & 5 PCs (100\% EV)                                                                & 0.9963 & 0.0559 & 0.9963 & 0.0560 & 0.9955 & 0.0621 \\ \hline
PCR2   & PCR     & \begin{tabular}[c]{@{}l@{}}poly09\tnote{a}\\ mean-centering\end{tabular}      & 5 PCs (97\% EV)     & 0.9966 & 0.0543 & 0.9965 & 0.0544 & 0.9968 & 0.0520 \\  \hline
PCR3   & PCR     & \begin{tabular}[c]{@{}l@{}}poly09\tnote{a}\\ autoscaling\end{tabular}      & 10 PCs (79\% EV)        & 0.9884 & 0.0994 & 0.9877 & 0.1024 & 0.9880 & 0.1014 \\ \hline
PCR4   & PCR     & \begin{tabular}[c]{@{}l@{}}poly09\tnote{a}\\ vector\tnote{b}\end{tabular}   & 6 PCs (100\% EV)               & 0.9951 & 0.0644 & 0.9951 & 0.0646 & 0.9805 & 0.1293 \\ \hline
PCR5   & PCR     & \begin{tabular}[c]{@{}l@{}}poly09\tnote{a} \\ vector\tnote{b}\\ mean-centering\end{tabular} & 6 PCs (91\% EV) & 0.9951 & 0.0644 & 0.9951 & 0.0646 & 0.9802 & 0.1301 \\ \hline
PCR6   & PCR     & -      & 6 PCs (100\% EV)                                                                   & 0.9967 & 0.0534 & 0.9967 & 0.0535 & 0.9954 & 0.0629 \\ \hline
PCR7   & PCR     & mean-centering       & 6 PCs (100\% EV)                                                     & 0.9969 & 0.0518 & 0.9969 & 0.0519 & 0.9967 & 0.0531 \\ \hline
PCR8   & PCR     & autoscaling        & 5 PCs (100\% EV)                                                       & 0.9970 & 0.0506 & 0.9970 & 0.0508 & 0.9923 & 0.0811 \\ \hline
PLSR1  & PLSR    & poly09\tnote{a}     &4 LVs (100\% EV)                                                               & 0.9962 & 0.0567 & 0.9962 & 0.0568 & 0.9938 & 0.0730 \\ \hline
PLSR2  & PLSR    & \begin{tabular}[c]{@{}l@{}}poly09\tnote{a}\\ mean-centering\end{tabular}  & 5 LVs (100\% EV)          & 0.9968 & 0.0520 & 0.9968 & 0.0526 & 0.9948 & 0.0665 \\  \hline
PLSR3  & PLSR    & \begin{tabular}[c]{@{}l@{}}poly09\tnote{a}\\ autoscaling\end{tabular}   & 10 LVs (78\% EV)           & 0.9983 & 0.0378 & 0.9957 & 0.0606 & 0.9942 & 0.0704 \\ \hline
PLSR4  & PLSR    & \begin{tabular}[c]{@{}l@{}}poly09\tnote{a}\\ vector\tnote{b}\end{tabular}     & 6 LVs (100\% EV)              & 0.9960 & 0.0587 & 0.9957 & 0.0603 & 0.9774 & 0.1389 \\ \hline
PLSR5  & PLSR    & \begin{tabular}[c]{@{}l@{}}poly09\tnote{a} \\ vector\tnote{b}\\ mean-centering\end{tabular} & 6 LVs (91\% EV) & 0.9972 & 0.0494 & 0.9963 & 0.0563 & 0.9760 & 0.1433 \\ \hline
PLSR6  & PLSR    & -   & 5 LVs (100\% EV)                                                                       & 0.9962 & 0.0570 & 0.9962 & 0.0571 & 0.9901 & 0.0921 \\ \hline
PLSR7  & PLSR    & mean-centering   & 5 LVs (100\% EV)                                                         & 0.9969 & 0.0514 & 0.9969 & 0.0516 & 0.9970 & 0.0510 \\ \hline
PLSR8  & PLSR    & autoscaling   & 5 LVs (100\% EV)                                                            & 0.9971 & 0.0501 & 0.9970 & 0.0504 & 0.9930 & 0.0773 \\ \hline
Ridge1 & Ridge   & k = 0.1     &                                                              & 0.9983 & 0.0384 & 0.9976 & 0.0456 & 0.9950 & 0.0656 \\ \hline
Ridge2 & Ridge   & k = 0.2      &                                                             & 0.9979 & 0.0420 & 0.9974 & 0.0468 & 0.9946 & 0.0678 \\ \hline
Ridge3 & Ridge   & k = 0.3       &                                                            & 0.9977 & 0.0441 & 0.9973 & 0.0480 & 0.9942 & 0.0704 \\
    \hline
    \end{tabular}
    }
      \begin{tablenotes}
        \item[a]Polynomial baseline correction (order = 9)
        \item[b]Vector normalisation
        \item[c]PC: principal component, LV: latent variable, EV: cumulative explained variance.
      \end{tablenotes}
\end{threeparttable}
\label{tbl:methanol_regression}
\end{table}

\begin{table}[ht]
\caption{Brand identification accuracy using through-bottle spectra}
\label{tab:Through-bottle}
  \begin{threeparttable}
  \resizebox{\textwidth}{!}{
  \renewcommand{\arraystretch}{1.2}
\begin{tabular}{lccccc}
\hline
            & \multicolumn{5}{c}{Accuracy (\%)}                                 \\ \cline{2-6} 
            & VV\tnote{a} & TT\tnote{b} & VT\tnote{c}& TV\tnote{d}& Mix\tnote{e}\\ \hline
KNN         & 100.0/100.0 & 100.0/100.0 & 100.0/37.0 & 100.0/21.3 & 99.8/100.0  \\ \hline
LDA         & 99.5/92.2   & 98.1/98.9   & 100.0/30.0 & 99.7/40.0  & 98.8/77.2   \\ \hline
ANN         & 36.2/20.0   & 61.9/51.1   & 20.0/24.0  & 41.3/30.0  & 42.1/34.4   \\ \hline
RF          & 100.0/100.0 & 100.0/98.9  & 100.0/39.7 & 100.0/20.0 & 100.0/97.8  \\ \hline
RBF SVM     & 61.0/57.8   & 45.7/31.1   & 60.0/30.0  & 61.3/20.0  & 57.4/47.8   \\ \hline
PCA+KNN     & 100.0/100.0 & 100.0/100.0 & 100.0/37.0 & 100.0/21.3 & 99.8/100.0  \\ \hline
PCA+LDA     & 100.0/100.0 & 90.5/86.7   & 100.0/19.0 & 88.7/30.0  & 74.3/72.8   \\ \hline
PCA+ANN     & 100.0/98.9  & 50.5/45.6   & 100.0/38.0 & 100.0/52.3 & 96.2/97.2   \\ \hline
PCA+RF      & 100.0/100.0 & 100.0/100.0 & 100.0/28.0 & 100.0/26.0 & 100.0/100.0 \\ \hline
PCA+RBF SVM & 61.0/57.8   & 46.2/32.2   & 60.0/30.7  & 62.3/20.0  & 66.0/58.9   \\ \hline
\end{tabular}%
}
      \begin{tablenotes}
        \item[a]The model was trained and tested using measurements obtained through vials (V).
        \item[b]The model was trained and tested using measurements obtained through bottles (T).
        \item[c]The model was trained using measurements obtained through vials (V) and tested using measurements obtained through bottles (T).
        \item[d]The model was trained using measurements obtained through bottles (T) and tested using measurements obtained through vials (V).
        \item[e]The training and test datasets include measurements obtained through both vials and bottles.
        \item The accuracy results are presented in the format of "Accuracy on training set/Accuracy on test set"
      \end{tablenotes}
\end{threeparttable}
\end{table}

\end{document}